\documentclass[journal]{IEEEtran}
\usepackage{amsmath,amsfonts}
\usepackage{algorithmic}
\usepackage{algorithm}
\usepackage{array}
\usepackage[caption=false,font=normalsize,labelfont=sf,textfont=sf]{subfig}
\usepackage{textcomp}
\usepackage{stfloats}
\usepackage{url}
\usepackage{verbatim}
\usepackage{graphicx}
\usepackage{cite}
\usepackage{tabularx}

\usepackage[utf8]{inputenc} % allow utf-8 input
\usepackage[T1]{fontenc}    % use 8-bit T1 fonts
\usepackage{hyperref}       % hyperlinks
\hypersetup{colorlinks=true,linkcolor=blue,citecolor=blue,urlcolor=blue}
\usepackage{booktabs}       % professional-quality tables
\usepackage{nicefrac}       % compact symbols for 1/2, etc.
\usepackage{microtype}      % microtypography
\usepackage{xcolor}         % colors

\usepackage{latexsym}
\usepackage{multirow}
\usepackage{bm}
\usepackage{graphicx}
\usepackage{wrapfig}
\usepackage{amssymb}
\usepackage{amsmath}
\usepackage{amsthm}
\usepackage{mathrsfs}
\usepackage{makecell}
\usepackage{colortbl}
\usepackage{tabularx}
\usepackage[normalem]{ulem}
\usepackage{pifont}

\definecolor{GrassGreen}{RGB}{146,208,80}
\definecolor{SkyBlue}{RGB}{4,175,252}
\definecolor{FluoGreen}{RGB}{24,229,25}
\definecolor{Fig1aGreen}{RGB}{106, 234, 102}
\definecolor{Fig1aBlue}{RGB}{0, 176, 240}
\definecolor{EncoderBlue}{RGB}{166, 203, 220}
\definecolor{DecoderGreen}{RGB}{131, 186, 158}
\definecolor{Pending}{RGB}{255, 0, 0}
\definecolor{red}{RGB}{255, 0, 0}
\definecolor{green}{RGB}{0, 255, 0}
\definecolor{blue}{RGB}{0, 0, 255}
\definecolor{AblationRed}{RGB}{251, 221, 238}
\definecolor{AblationRedWord}{RGB}{231, 146, 238}
\definecolor{AblationGreen}{RGB}{193, 220, 206}
\definecolor{AblationGreenWord}{RGB}{160, 191, 118}
\usepackage{xcolor}
\usepackage{capt-of}
\usepackage[normalem]{ulem}

\title{XS-VID: A Large-Scale Benchmark for Small Object Detection and Tracking in Videos}

\author{Jiahao Guo, Ziyang Xu, Lianjun Wu, Fei Gao, Wenyu Liu, and Xinggang Wang%
\IEEEcompsocitemizethanks{
\IEEEcompsocthanksitem{The authors are with the School of Electronic Information and Communications, Huazhong University of Science and Technology, Wuhan 430074, China.}
\IEEEcompsocthanksitem{Corresponding author: Xinggang Wang (email: \url{xgwang@hust.edu.cn}).}
}}

\makeatletter
\g@addto@macro\@maketitle{
\begin{center}
    \setcounter{figure}{0}
    \includegraphics[width=\linewidth]{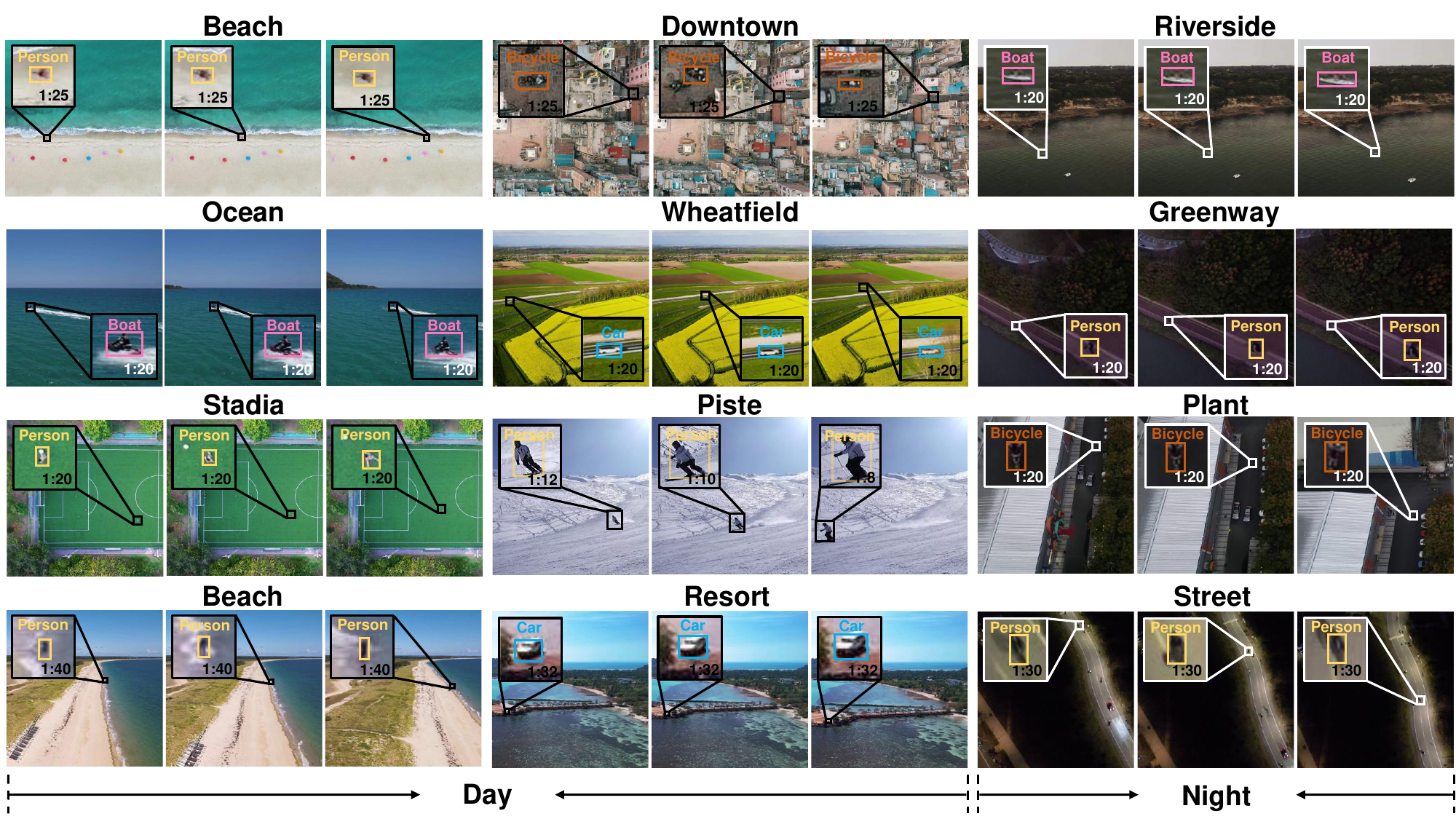}
    \captionof{figure}{\textbf{Extremely small objects across diverse scenes in XS-VID.} The dataset features challenging small-object detection scenarios with scale ratios as extreme as 1:40, spanning 16 scene types (11 illustrated here) including natural environments, urban areas, recreational zones, and infrastructure sites under day and night conditions. Bounding boxes highlight the minute scale of annotated objects, demonstrating XS-VID's focus on extremely small-object detection challenges.}
    \label{fig:showcases}
\end{center}
\vspace{-10mm}
}
\makeatother

\begin{document}
\maketitle

\begin{abstract}
Small object detection and tracking in videos remain critical yet underexplored challenges in computer vision, particularly for applications such as public safety, aerial surveillance, and autonomous driving. Existing benchmarks offer limited support due to limited numbers of small objects, constrained category diversity, and narrow scene coverage. To address these limitations, we introduce \textbf{XS-VID}, a large-scale video benchmark comprising 223K frames and 1.4M annotated bounding boxes across 374 video sequences spanning diverse scene types. XS-VID provides extensive coverage of small-object scales, particularly for extremely small ($0\sim12^2$ pixels) and small ($12^2\sim20^2$ pixels) objects, which collectively constitute over 55\% of all annotations. For systematic evaluation, we establish three dedicated tracks: Detection, multiple object tracking (MOT), and single object tracking (SOT), and extensively test the existing state-of-the-art methods on each. The experimental results indicate that existing methods face significant challenges with XS-VID, mainly stemming from insufficient modeling of spatiotemporal features at small scales. To tackle these challenges, we propose a lightweight, high-precision detection framework dubbed \textbf{YOLOFT}. It enhances small-object feature representation and spatiotemporal integration while preserving high detection speed, thereby achieving improved accuracy and robustness on both the XS-VID and VisDrone benchmarks. Our dataset and code are publicly available at \url{https://gjhhust.github.io/XS-VID/}, providing a solid foundation for future research on small-object detection and tracking in videos.
\end{abstract}


\begin{figure}[htbp]
    \begin{minipage}[t]{0.48\textwidth}
        \centering
        \includegraphics[width=\linewidth]{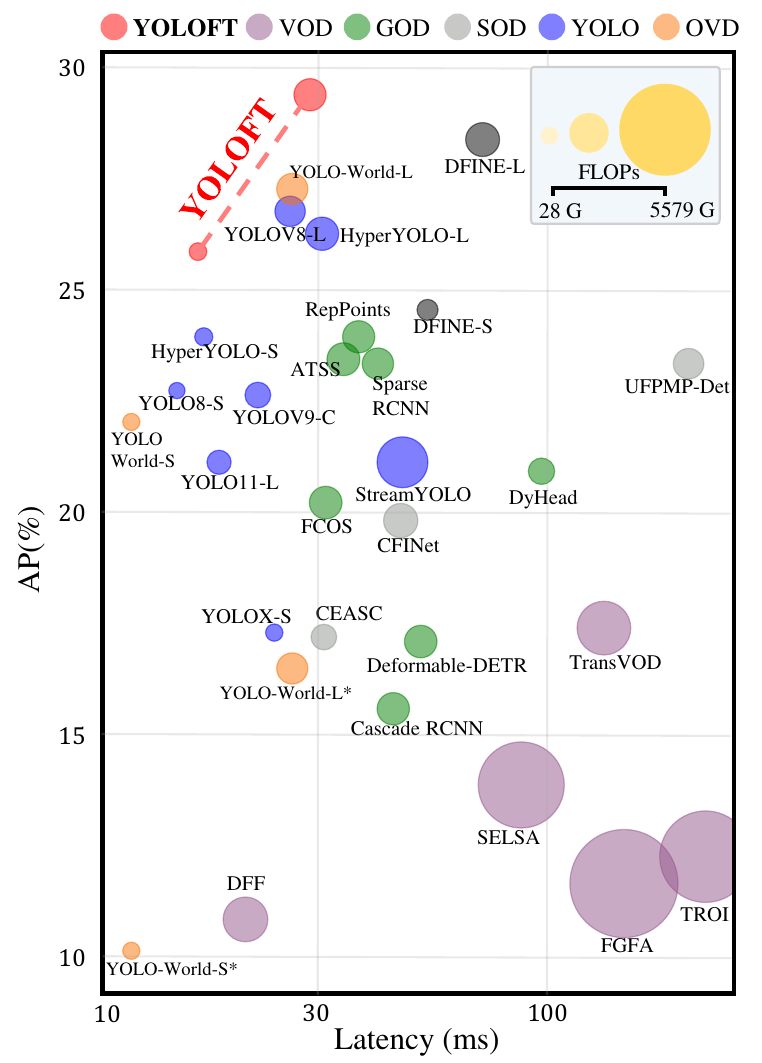}
        \caption{\textbf{AP-Latency comparison of various methods on our XS-VID.} Our YOLOFT achieves the SOTA performance. Latency calculations are all with batch size = 1. Details in Tab.~\ref{tab:xsvid_det_results}.}
        \label{fig:bubble_model}
    \end{minipage}%
\vspace{-10pt}

\end{figure}

\section{Introduction}

\begin{figure*}
\centering
\includegraphics[width=1.0\linewidth]{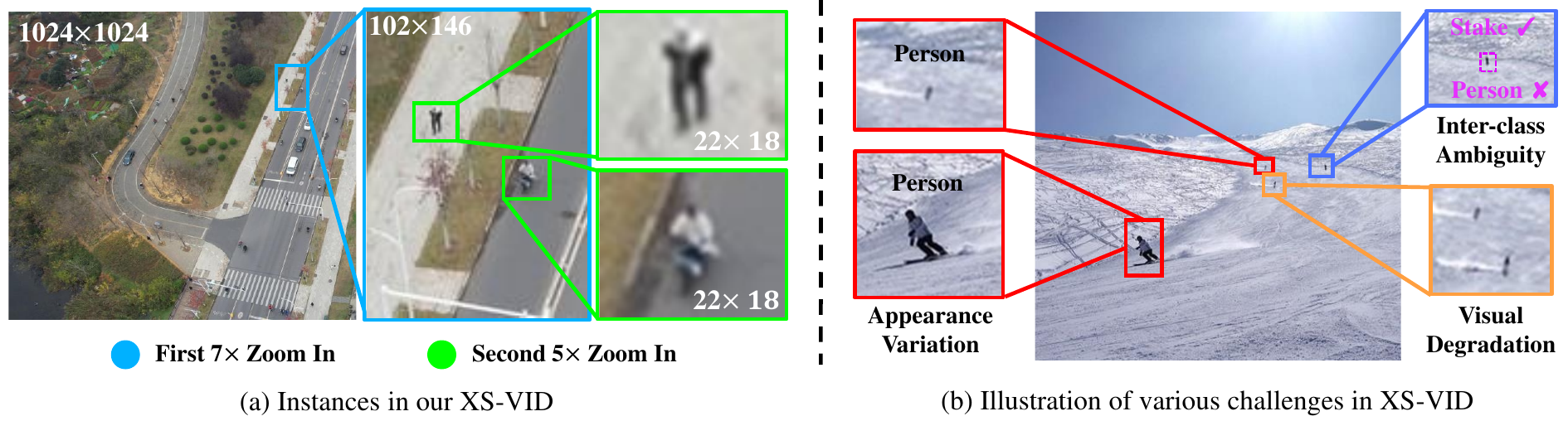}
\caption{\textbf{Showcases of our XS-VID dataset's object size and challenges in small-object detection.} (a) shows that the objects in our XS-VID dataset are extremely small, and (b) indicates that small-object detection mainly faces three challenges: high appearance variation (multi-appearance), semantic ambiguity between classes (mis-classification), and severe visual degradation caused by low resolution and insufficient texture details (texture distortion).}
\label{fig:challenge_SVOD}
\vspace{-5pt}
\end{figure*}

\IEEEPARstart{S}{mall} objects pose fundamental challenges in video-based tasks such as video object detection (VOD), single object tracking (SOT), and multi-object tracking (MOT), especially in real-world applications like public safety, aerial surveillance, and autonomous driving \cite{UAVDT, zhu2021detectionVisdrone, Cheng_2023SODA, RGBT-Tiny}. Despite recent advances in object detection and tracking methods, detecting small objects in videos remains inadequately addressed. This oversight is primarily due to the lack of video datasets specifically designed for small-object detection and tracking~\cite{LSTFE}, which hinders both training and evaluation processes, and consequently leads to a scarcity of effective methods.

Several recent efforts have explored small-object detection from different perspectives. For instance, VisDrone~\cite{zhu2021detectionVisdrone} focuses on UAV-based detection in traffic-heavy scenarios, SODA~\cite{Cheng_2023SODA} targets high-altitude aerial remote sensing scenes, and RGBT-Tiny~\cite{RGBT-Tiny} investigates tiny object detection in dual-modality (RGB and thermal) settings. While valuable, these datasets and surveys often emphasize modality-specific or scene-specific challenges, lacking a unified foundation for studying small objects across general-purpose video settings. To complement these efforts, we conduct a more systematic investigation focused on the fundamental challenges of detecting and tracking small objects under diverse conditions.

Moreover, existing mainstream video datasets, such as ImageNetVID~\cite{russakovsky2015imagevid}, VisDrone~\cite{zhu2021detectionVisdrone}, and UAVDT~\cite{UAVDT}, contain few small objects. As shown in Tab.~\ref{tab:dataset_detail}, objects smaller than $12^2$ pixels account for merely 0.1\%, 0.5\%, and 3.3\% of all the annotations, respectively. These datasets also exhibit structural limitations: UAVDT is restricted to vehicle categories in traffic scenes, VisDrone excludes far-field pedestrians from annotation, and ImageNetVID primarily includes medium-to-large objects. Such biases hinder both the development and fair evaluation of small-object detection and tracking algorithms.

To overcome these limitations, we introduce \textbf{XS-VID}, a large-scale video benchmark specifically designed for small-object detection and tracking. As illustrated in Fig.~\ref{fig:showcases}, XS-VID contains 223K frames and 374 video sequences annotated with 1.4M bounding boxes, spanning 16 diverse scene types such as forest, mountain ranges, airports, city roads, and snowy fields, across various times of day and night. It encompasses a wide range of object scales, with a particular emphasis on small objects. The small objects in XS-VID are not concentrated at a fixed size but comprehensively cover small-object sizes ranging from $0\sim32^2$. Specifically, the numbers of objects within the ranges of \textit{es} (extremely small; $0\sim12^2$), \textit{rs} (relatively small; $12^2\sim20^2$), \textit{gs} (generally small; $20^2\sim32^2$), and \textit{normal} (> $32^2$) are 418k, 370k, 237k, and 394k, respectively, accounting for 29.4\%, 26.1\%, 16.7\%, and 27.8\%. To our best knowledge (details in Tab.~\ref{tab:dataset_summary}), XS-VID offers the most comprehensive coverage of small-object sizes, the largest number and proportion of \textit{es} and \textit{rs} objects, and the widest range of scene types, effectively addressing limitations in existing benchmarks.

For targeted evaluation, XS-VID introduces three tracks: Detection, Single Object Tracking (SOT), and Multi-Object Tracking (MOT). Each track evaluates specific capabilities of detection and tracking models in small-object video settings. Empirical results show that existing methods consistently underperform in small-object video detection, particularly under the challenging conditions represented in XS-VID. These limitations highlight algorithmic deficiencies that have received limited attention, largely due to the absence of dedicated large-scale benchmarks in this area. As illustrated in Fig.~\ref{fig:challenge_SVOD}, XS-VID exposes key challenges such as high appearance variation, inter-class ambiguity, and visual degradation caused by low resolution and insufficient textures.

To explore effective solutions for small-object detection and tracking, we conduct extensive evaluations across all three tracks using a variety of state-of-the-art (SOTA) methods. Our analysis reveals three major limitations that affect small-object performance: weak feature extraction due to insufficient scale-aware modeling, inadequate localization supervision for extremely small instances, and ineffective spatial reconstruction during the upsampling process. Furthermore, most existing approaches either ignore temporal information or fail to utilize it effectively. Motivated by these observations, we enhance the YOLO framework by incorporating \textbf{F}low-guided feature enhancement~\cite{SlowFast, Girdhar_2019_CVPR, teed2020raft} and \textbf{T}emporal modeling~\cite{I3D, X3D}, leading to our proposed architecture, \textbf{YOLOFT}. By integrating a Flow Mask Attention module, YOLOFT improves spatiotemporal consistency and localization precision while maintaining a lightweight design and computational efficiency. Extensive experiments demonstrate that YOLOFT achieves strong performance on both XS-VID and VisDrone.

In summary, our main contributions are as follows:

(1) We present XS-VID, a large-scale video dataset with \textbf{1.4 million} bounding box annotations\text{-}comparable in scale to ImageNetVID\text{-}that specifically targets small-object detection and tracking. XS-VID covers the most comprehensive range of small-object sizes to date, with \textbf{72.2\%} (\textbf{1025K} instances) of annotations smaller than $32^2$ pixels and \textbf{29.4\%} (\textbf{418K} instances) classified as extremely small ($<12^2$ pixels). It also includes \textbf{16} diverse scene types, substantially surpassing existing benchmarks in object scale (418K vs. <30K small objects) and scene diversity .

(2) We perform a comprehensive survey of \textbf{30} small-object detection datasets and \textbf{46} representative methods, identifying key limitations and design bottlenecks that inform the construction and evaluation protocol of XS-VID.

(3) We establish three dedicated evaluation tracks\text{-}Detection, Multi-Object Tracking (MOT), and Single-Object Tracking (SOT)\text{-}to enable targeted analysis of detection and tracking performance in small-object scenarios. Extensive experiments show that SOTA methods struggle significantly under the challenging conditions of XS-VID.

(4) We propose YOLOFT as a lightweight video detection baseline tailored for small objects. It integrates dynamic temporal modeling and flow-guided attention mechanisms to enhance spatiotemporal consistency and fine-grained localization. YOLOFT achieves \textbf{29.3\%} AP(\textbf{+1.0\%}) on XS-VID and best AP$_{es}$(\textbf{+2.3\%}) on VisDrone, providing a practical and efficient baseline for future research.

\section{Related Work}
\subsection{Datasets for Small Object Detection and Tracking}
Small-object detection faces challenges from limited resolution, weak texture, and frequent occlusion. Previous work addresses these through specialized architectures and custom benchmarks~\cite{Cheng_2023SODA, RGBT-Tiny}.
Several datasets have been proposed to support small-object research.
In the RGB domain, the SODA family (SODA-D and SODA-A) focuses on driving and aerial scenes with dense annotations, while SODA-10M provides a large-scale benchmark for self-supervised and multi-task pretraining~\cite{SODA10M}. SOD4SB~\cite{SOD4SB} covers tiny bird detection with over 137K annotations. Other datasets like TJU-DHD Traffic~\cite{TJU-DHDTraffic}, WiderFace~\cite{WiderFace}, WiderPerson~\cite{WiderPerson}, and EuroCity Persons~\cite{EuroCityPersons} include many small objects in human or traffic scenes. In the aerial domain, DOTA v2~\cite{DOTA} offers multi-resolution overhead imagery that has become standard for remote-sensing small-object detection.
For thermal and multimodal sensing, SIRST-v2~\cite{OSCAR}, IRSTD-1K~\cite{ISNet}, and NUDT-SIRST~\cite{DNAnet} focus on infrared detection of single-class distant or dim targets. RGB-T datasets such as DroneVehicle~\cite{UA-CMDet}, RGBTDronePerson~\cite{QFDet}, VEDAI~\cite{VEDAI}, and RGBT-Tiny~\cite{RGBT-Tiny} support multi-modal fusion under low-light or occlusion. Tracking datasets like Anti-UAV410~\cite{Anti-UAV410} and VTUAV~\cite{VTUAV} provide long video sequences for UAV tracking in challenging conditions.
While these datasets span diverse modalities and scenarios, they predominantly target static images, single-class detection, or non-temporal tasks.

\begin{table}[t]
  \centering
  \caption{Comparison of Small Object Area Distribution Across Various Datasets.}
  \label{tab:dataset_detail}
  \setlength{\tabcolsep}{3pt}
  \renewcommand{\arraystretch}{0.9}
  \resizebox{\columnwidth}{!}{%
  \begin{tabular}{lrrrrrr}
    \toprule
    \multirow{2}{*}{Dataset} & \multicolumn{2}{r}{$0\sim 12^2$ (\textit{es})} & \multicolumn{2}{r}{$12^2\sim 20^2$ (\textit{rs})} & \multicolumn{2}{r}{$20^2\sim 32^2$ (\textit{gs})} \\
    \cmidrule{2-7}
    & Count & Ratio & Count & Ratio & Count & Ratio \\
    \midrule
    ImageNetVID~\cite{russakovsky2015imagevid} & 968 & 0.1\% & 13k & 0.7\% & 47k & 2.4\% \\
    UAVDT~\cite{UAVDT} & 29k & 3.3\% & 204k & 23.0\% & 265k & 29.8\% \\
    VisDrone VID~\cite{zhu2021detectionVisdrone} & 7k & 0.5\% & 115k & 7.5\% & 316k & 20.6\% \\
    XS-VID (ours) & \textbf{418k} & \textbf{29.4\%} & \textbf{370k} & \textbf{26.1\%} & 237k & 16.7\% \\
    \bottomrule
  \end{tabular}}
\end{table}

\begin{table*}[htbp]
    \centering
    \caption{
    \textbf{Comprehensive comparison of representative benchmarks for object detection, tracking, and joint detection and tracking (D\&T) under small object scenarios.} The \textbf{Type} column indicates the sensing modality, including RGB (visible), IR (infrared), and IR+RGB (dual-modality). \textbf{T-Cat.} denotes the number of target categories, and \textbf{S-Cat.} denotes the number of distinct scene types. Datasets are grouped by sensing modality (Type). Resolution denotes the released image size or the maximum image size for datasets with varying resolutions.
    XS-VID features \textbf{1.4 million annotations}, which is comparable in scale to ImageNet VID (2 million), while focusing on extremely small objects. It combines extensive extremely small-object coverage with Detection, SOT, and MOT evaluation. Datasets marked with * are widely used but not specifically designed for small object detection. Varied means that the image sizes are not uniform.
    }
    \label{tab:dataset_summary}
    \resizebox{\linewidth}{!}{
    \begin{tabular}{lccclcccccccr}
    \toprule
    Dataset & Det. & Track. & Type & Domain & Seq. & Frames & Annotations & Resolution & FPS & T-Cat. & S-Cat. & Pub \& Year \\
    \midrule
    SIRST-v2~\cite{OSCAR} & \checkmark & & IR & infrared & - & 514 & 648 & 278$\times$366 & - & 1 & 3 & TGRS 2023 \\
    IRSTD-1K~\cite{ISNet} & \checkmark & & IR & infrared & - & 1K & 1.5K & 512$\times$512 & - & 1 & 6 & CVPR 2022 \\
    NUDT-SIRST~\cite{DNAnet} & \checkmark & & IR & infrared & - & 1.3K & 1.9K & 256$\times$256 & - & 1 & 5 & TIP 2022 \\
    IRDST~\cite{RDIAN} & \checkmark & & IR & infrared & 401 & 147K & 144K & 795$\times$553 & - & 1 & - & TGRS 2023 \\
    SIATD~\cite{SIATD} & \checkmark & & IR & infrared & 350 & 150K & 247K & 640$\times$512 & 30 & 1 & 3 & CSD 2021 \\
    Hui~\cite{Hui1} & \checkmark & \checkmark & IR & infrared & 22 & 16K & 17K & 256$\times$256 & 100 & 2 & - & CSD 2020 \\
    Fu~\cite{Hui2} & \checkmark & \checkmark & IR & infrared & 87 & 22K & 89K & 640$\times$512 & 35 & 1 & 2 & CSD 2022 \\
    \midrule
    RGBTDronePerson~\cite{QFDet} & \checkmark & & IR+RGB & UAV & - & 16.8K & 74K & 640$\times$512 & - & 1 & - & ISPRS 2023 \\
    DroneVehicle~\cite{UA-CMDet} & \checkmark & & IR+RGB & UAV & - & 57K & 953K & 840$\times$712 & - & 5 & - & TCSVT 2021 \\
    VEDAI~\cite{VEDAI} & \checkmark & & IR+RGB & aerial & - & 1.2K & 3.7K & Varied & - & 9 & - & JVCI 2016 \\
    RGBT-Tiny~\cite{RGBT-Tiny} & \checkmark & & IR+RGB & UAV & 115 & 93K & 1.2M & 640$\times$512 & 15 & 7 & 8 & TPAMI 2024 \\
    VTUAV~\cite{VTUAV} & & \checkmark & IR+RGB & UAV & 500 & 3.3M & 326K & 1920$\times$1080 & 30 & 13 & 15 & CVPR 2022 \\
    Anti-UAV410~\cite{Anti-UAV410} & & \checkmark & IR+RGB & UAV & 410 & 438K & 438K & 640$\times$512 & 25 & 1 & 7 & TPAMI 2024 \\
    \midrule
    ImageNetVID* ~\cite{russakovsky2015imagevid} & \checkmark & & RGB & general & 4417 & 1.2M & 2M & Varied & 25 & 30 & - & IJCV 2015 \\
    SOD4SB~\cite{SOD4SB} & \checkmark & & RGB & wildlife & - & 39K & 137K & 3840$\times$2160 & - & 4 & 4 & CVPR 2023 \\
    SODA-D~\cite{Cheng_2023SODA} & \checkmark & & RGB & aerial & - & 24K & 278K & 3407$\times$2470 & - & 9 & - & TPAMI 2023 \\
    SODA-A~\cite{Cheng_2023SODA} & \checkmark & & RGB & aerial & - & 2.5K & 800K & 4761$\times$2777 & - & 9 & - & TPAMI 2023 \\
    TJU-DHD Traffic~\cite{TJU-DHDTraffic} & \checkmark & & RGB & traffic & - & 60K & 332K & 1624$\times$1200 & - & 5 & - & TIP 2021 \\
    SODA-10M~\cite{SODA10M} & \checkmark & & RGB & aerial & - & 10K & 13K & 1920$\times$1080 & - & 6 & - & NeurIPS 2021 \\
    DOTA v2~\cite{DOTA} & \checkmark & & RGB & aerial & - & 11K & 1.8M & Varied & - & 18 & - & TPAMI 2021 \\
    WiderPerson~\cite{WiderPerson} & \checkmark & & RGB & person & - & 13K & 400K & Varied & - & 1 & - & TMM 2020 \\
    EuroCity Persons~\cite{EuroCityPersons} & \checkmark & & RGB & traffic & - & 47K & 211K & 1920$\times$1024 & - & 3 & - & TPAMI 2020 \\
    WiderFace~\cite{WiderFace} & \checkmark & & RGB & face & - & 32K & 394K & Varied & - & 1 & 61 & CVPR 2016 \\
    MS-COCO~\cite{COCO} & \checkmark & & RGB & general & - & 328K & 2.5M & Varied & - & 91 & - & ECCV 2014 \\
    FL-Drones~\cite{FL-Drones} & \checkmark & & RGB & UAV & 14 & 38K & 20K & Varied & -- & 1 & 1 & TPAMI 2016 \\
    UAVDL~\cite{Mueller2016UAV123} & & \checkmark & RGB & UAV & 20 & 58K & 58K & 1280$\times$720 & 30/96 & 5 & 1 & ECCV 2016 \\
    UAV123~\cite{Mueller2016UAV123} & & \checkmark & RGB & UAV & 123 & 106K & 106K & 1280$\times$720 & 30/96 & 7 & 3 & ECCV 2016 \\
    DarkTrack2021~\cite{DarkTrack2021} & & \checkmark & RGB & UAV & 110 & 100k & - & - & 30 & 5 & 1 & ICRA 2021 \\
    NPS-Drones~\cite{NPS-Drone} & \checkmark & \checkmark & RGB & UAV & 50 & 70K & 70K & Varied & 30 & 1 & 1 & IROS 2016 \\
    UAVDT~\cite{UAVDT} & \checkmark & \checkmark & RGB & traffic & 100 & 80K & 840K & 1080$\times$540 & 30 & 6 & 1 & ECCV 2018 \\
    VisDrone VID~\cite{zhu2021detectionVisdrone} & \checkmark & \checkmark & RGB & UAV & 40 & 183K & 183K & 756$\times$1344 & - & 10 & - & TPAMI 2021 \\
    \textbf{XS-VID} & \checkmark & \checkmark & RGB & \textbf{general} & \textbf{374} & 223K & \textbf{1.4M} & 1024$\times$1024 & 25 & 7 & \textbf{16} & - 2026 \\
    \bottomrule
    \end{tabular}
    }
\end{table*}

\subsection{Datasets for Video Object Detection and Tracking}
Compared to image detection, video detection exploits temporal redundancy and motion continuity to recover blurred or occluded objects~\cite{LSTFE}. However, most video benchmarks do not explicitly address the challenges of small-object detection.
ImageNetVID~\cite{russakovsky2015imagevid}, one of the earliest large-scale video detection benchmarks, contains over one million annotations across 30 categories. Yet, only about 3\% of objects are under $32^2$ pixels, making it poorly suited for tiny object evaluation. VisDrone~\cite{zhu2021detectionVisdrone} includes diverse urban and rural scenes captured by low-altitude UAVs and a larger share of small objects, yet still lacks extremely small instances. Its category set focuses mainly on traffic-related classes, limiting broader applicability.
UAVDT~\cite{UAVDT} emphasizes vehicle detection in drone videos and contains around 3.3\% of objects under $12^2$ pixels. However, it remains limited to a single class and lacks viewpoint diversity. Anti-UAV410~\cite{Anti-UAV410} provides RGB-T videos for anti-drone tracking but focuses solely on UAV targets. Recent studies~\cite{Dogfight, LSTFE} emphasize the lack of suitable video datasets for tiny object detection.
Overall, existing video benchmarks often lack scene diversity, broad category coverage, and sufficient small objects.

\begin{figure*}[!ht]
    \centering
    \includegraphics[width=1.0\linewidth]{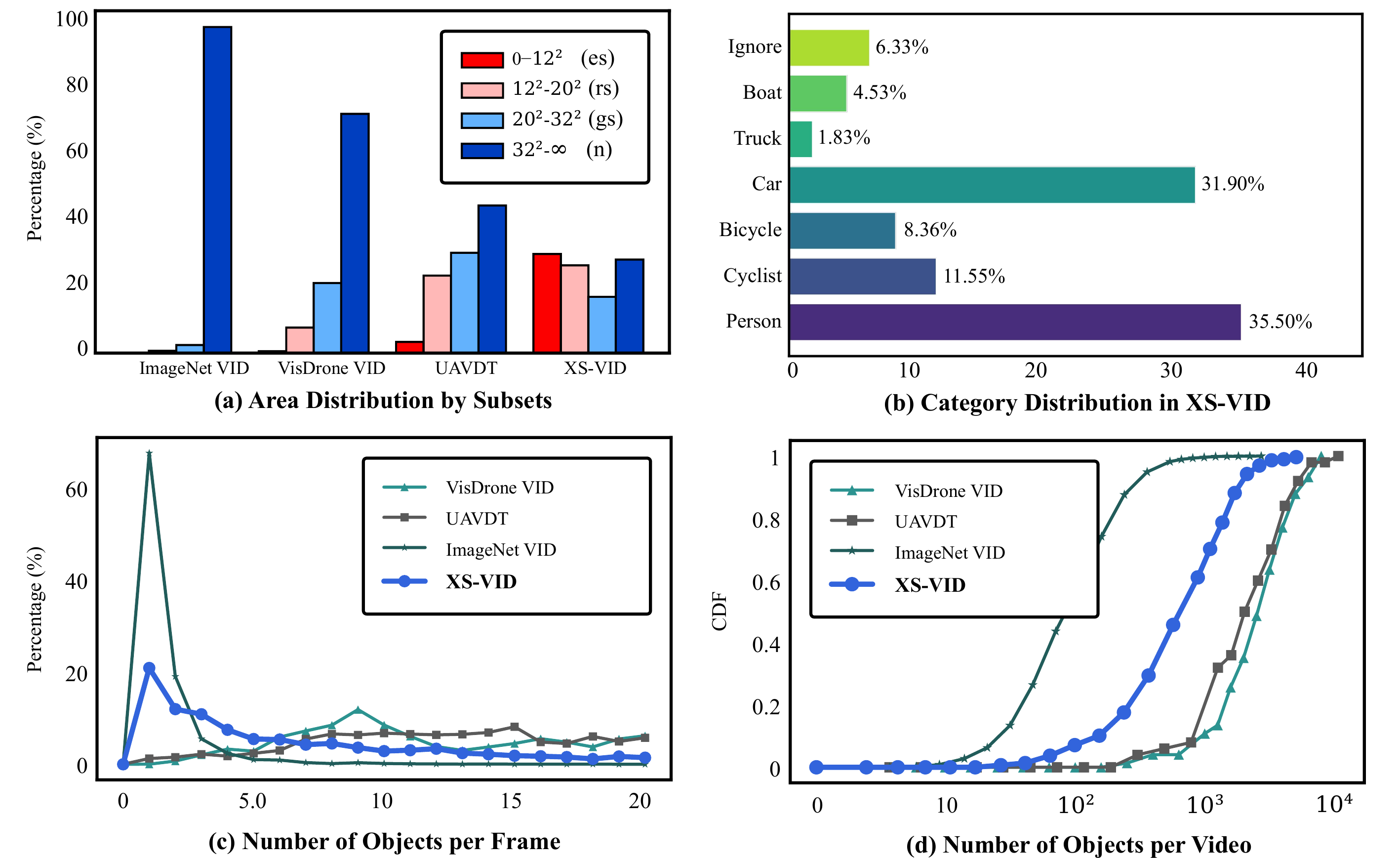}
    \caption{\textbf{Quantitative Comparison between XS-VID and various datasets.} (a) XS-VID exhibits a significantly higher proportion of extremely small ($0\!\sim\!12^2$) and relatively small ($12^2\!\sim\!20^2$) objects than other datasets.
    (b) Category distribution across XS-VID's 7 classes.
    (c) Number of objects per frame in XS-VID is rich and balanced, compared to ImageNetVID and VisDrone.
    (d) The cumulative distribution of total object counts per video shows XS-VID emphasizes scene diversity with moderate video total target number.}
    \label{fig:dataset-Statistics}
    % \vspace{-10pt}
\end{figure*}

\section{The XS-VID Dataset}

We developed XS-VID, a large-scale video dataset specifically designed for small-object detection and tracking, through rigorous screening and annotation processes. It comprises 374 video sequences with a resolution of $1024 \times 1024$, totaling 222{,}924 images and 1{,}419{,}009 annotated object boxes, each with detailed bounding box and trajectory annotations. The dataset is split into training and testing sets based on non-overlapping scenes, resulting in an approximate 5:1 ratio of training to testing samples. Detailed information on the dataset construction pipeline, including collection procedures, category definitions, annotation protocols, post-processing, and quality assurance strategies, is provided in the Supplementary Material (SM), which also includes representative examples from XS-VID illustrating object scale variation, scene diversity, and annotation richness.

\subsection{Data Collection and Annotation}

We collected 500 high-resolution videos ($3840 \times 2160$, 25\,fps) using DJI Air3 drones across a wide range of environments and conditions. From these, 204 high-quality clips were selected, and an additional 170 videos were
carefully curated from online sources to maximize scene diversity and coverage. The dataset covers seven categories: person, car, bicycle, cyclist, boat, truck, and ignore. To ensure privacy protection, we carefully cropped or blurred all sensitive regions including faces, license plates, and identifiable building features throughout the dataset. Annotating small instances—particularly pedestrians—is challenging due to frequent occlusion and low texture.

Annotations were generated via a semi-automatic pipeline, starting with CO-DETR~\cite{codetr2022} for initial predictions, followed by interval-based manual refinement. Identity assignment was performed using feature interpolation~\cite{lowe2004distinctiveSIFT, muja2009FLANN}, and final annotations were verified by multiple annotators. All frames were labeled at the original $3840 \times 2160$ resolution to ensure high accuracy, especially for extremely small objects. For deployment-oriented evaluation, videos were uniformly downsampled by a factor of two to $1920 \times 1080$, balancing annotation precision with practical resolution settings. The square $1024\times1024$ release format and interpolation details are provided in the SM's Sec.~I-A. Extensive quality control was applied, with inter-annotator agreement (IoU) exceeding 0.85 on a test subset.

\subsection{Statistical Analysis}

We begin by summarizing representative benchmarks for small-object detection and tracking in Tab.~\ref{tab:dataset_summary}, covering a broad spectrum of RGB, infrared, and dual-modality datasets across detection and tracking tasks. While recent efforts such as RGBT-Tiny, SODA, and UAVDT have addressed small-object scenarios in specific modalities or domains, most lack the scale, resolution, and scene diversity required for general-purpose video applications. In contrast, XS-VID provides a large-scale RGB benchmark with 223K frames, 1.4M annotated bounding boxes, and 374 video sequences spanning 16 diverse scenes, offering rich coverage for small-object research.

To provide a more detailed comparison, Tab.~\ref{tab:dataset_detail} analyzes small-object distribution across datasets, revealing XS-VID's exceptional focus on extremely small targets. XS-VID contains 418k objects in the most challenging size range ($0\sim 12^2$ pixels), representing 29.4\% of all annotations compared to ImageNetVID (0.1\%), UAVDT (3.3\%), and VisDrone (0.5\%). In addition, we present visual comparisons in Fig.~\ref{fig:dataset-Statistics}, including distributions of object sizes, number of objects per frame, and number of objects per video, highlighting XS-VID’s significantly higher density and proportion of extremely small objects compared to VisDrone, UAVDT, and ImageNetVID.

\subsubsection{Highest number of extremely small objects and the widest area distribution} Fig.~\ref{fig:dataset-Statistics}(a) shows that the proportion of objects in the size range of $0\sim12^2$ and $12^2\sim20^2$ in XS-VID is significantly higher than in other datasets. Objects in the $0\sim12^2$ range account for up to 29.4\% (compared to 0.5\%, 3.3\%, and 0.1\% for VisDrone, UAVDT, and ImageNetVID, respectively). Fig.~\ref{fig:dataset-Statistics}(b) shows the category distribution of XS-VID, containing seven categories.

\subsubsection{Rich and balanced number of objects per frame} As shown in Fig.~\ref{fig:dataset-Statistics}(c), the number of objects per frame in XS-VID is relatively balanced, whereas ImageNetVID focuses on large single objects per frame. Fig.~\ref{fig:dataset-Statistics}(d) shows the cumulative distribution of total object counts per video. ImageNetVID videos contain fewer than 100 total objects due to single large-object scenarios. UAVDT and VisDrone reach 4,000-5,000 objects per video due to longer sequences, while XS-VID emphasizes scene diversity, maintaining up to 1,000 objects per video across diverse scenarios.

\section{DETECTION TRACK}\label{sec:VOD_TRACK}

\begin{table*}[t]
\centering
\caption{Comparison of representative image- and video-based detection methods on the XS-VID Detection Track. "\#Params." refers to the number of model parameters. Method categories include VOD (video object detection), GOD (general object detection), SOD (small object detection), YOLO (YOLO series and variants), OVD (open-vocabulary detection), Unified (unified frameworks), and OURS (our method tailored for small object video detection, detailed in Section~\ref{section:YOLOFT}). Temporal and image-based methods share the same frame-level evaluation protocol. We report the configurations listed in the Method column using their official implementations.  ZS denotes zero-shot evaluation, and FT denotes fine-tuning on XS-VID. Reproduction details are provided in the SM's Secs.~II-A--II-B.}
\label{tab:xsvid_det_results}
\setlength{\tabcolsep}{5.2pt}
\begin{tabular}{llrrrrrrrrr}
\toprule
\multicolumn{2}{l}{Method} & $\text{AP}^{test}$ & $\text{AP}^{test}_{es}$ & $\text{AP}^{test}_{rs}$ & $\text{AP}^{test}_{gs}$ & $\text{AP}^{test}_{m}$ & $\text{AP}^{test}_{l}$ & \#Params. & FLOPs & Latency$\scriptstyle [bs=1]$ \\
\midrule
\multirow{8}{*}{\rotatebox{90}{VOD}} 
& DFF~\cite{Zhu_2017_CVPRdff}           & 10.9\% & 0.0\% & 0.2\% & 1.1\% & 7.6\%  & 23.0\% & 62.12 M  & 187.24 G & 20.9 ms \\
& FGFA~\cite{Zhu_2017_ICCVfgfa}         & 11.7\% & 0.0\% & 0.2\% & 1.5\% & 9.3\%  & 23.4\% & 64.48 M  & 5579.10 G & 151.5 ms \\
& SELSA~\cite{wu2019selsa}              & 13.9\% & 0.0\% & 0.5\% & 2.3\% & 11.6\% & 28.1\% & 70.63 M  & 2846.46 G & 88.5 ms \\
& TROI~\cite{TROI_2021}                 & 12.3\% & 0.0\% & 0.3\% & 1.8\% & 10.4\% & 25.6\% & 78.12 M  & 3416.44 G & 232.1 ms \\
& MEGA~\cite{Chen_2020_CVPRMEGA}        &  9.8\% & 0.0\% & 0.1\% & 0.6\% & 6.8\%  & 22.8\% & --       & --        & --      \\
& DiffusionVID~\cite{diffusionvid}      & 11.5\% & 0.0\% & 0.2\% & 1.4\% & 9.0\%  & 23.3\% & --       & --        & --      \\
& StreamYOLO~\cite{yang2022realtimeStreamYOLO} & 21.1\% & 7.2\% & 10.3\% & 11.1\% & 22.0\% & 35.5\% & 54.84 M  & 698.08 G  & 47.54 ms \\
& TransVOD~\cite{9960850TransVOD}       & 17.4\% & 3.8\% & 6.4\% & 6.0\% & 18.3\% & 30.1\% & 33.78 M  & 613.92 G  & 136.4 ms \\
\midrule
\multirow{10}{*}{\rotatebox{90}{GOD}} 
& FCOS~\cite{tian2019fcos}              & 20.2\% & 5.3\% & 7.8\% & 7.2\% & 20.0\% & 35.5\% & 32.1 M   & 202.0 G   & 31.8 ms \\
& ATSS~\cite{zhang2020atss}             & 23.4\% & 7.7\% & 10.9\%& 11.8\%& 25.3\% & 36.0\% & 32.1 M   & 206.0 G   & 34.9 ms \\
& DyHead~\cite{dai2021dynamichead}      & 20.9\% & 4.8\% & 8.2\% & 8.2\% & 21.6\% & 32.1\% & 38.906 M & 112.0 G   & 98.4 ms \\
& RepPoints~\cite{chen2020reppoints}    & 23.9\% & 6.2\% & 10.3\%& 11.0\%& 24.8\% & 41.3\% & 36.8 M   & 194.0 G   & 37.8 ms \\
& Deformable-DETR~\cite{zhu2020deformabledetr} & 17.1\% & 3.8\% & 5.2\% & 5.8\% & 16.5\% & 42.5\% & 40.1 M   & 200.0 G   & 52.3 ms \\
& Sparse RCNN~\cite{sun2021sparsercnn}  & 23.3\% & 7.2\% & 11.3\%& 12.3\%& 23.4\% & 39.4\% & 106.0 M  & 181.3 G   & 41.8 ms \\
& Cascade RCNN~\cite{Cascade_RCNN}      & 15.6\% & 1.9\% & 5.3\% & 5.5\% & 15.5\% & 23.0\% & 41.38 M  & 196.73 G  & 45.3 ms \\
& D-FINE-S~\cite{DFINE2025}             & 24.5\% & 9.6\% & 13.5\%& 12.2\%& 26.5\% & 35.5\% & 10.18 M  & 59.3 G    & 54.2 ms \\
& D-FINE-L~\cite{DFINE2025}             & 28.3\% & 11.0\%& 15.4\%& 15.1\%& 28.6\% & 48.3\% & 30.64 M  & 223.9 G   & 72.3 ms \\
& CO-DETR~\cite{codetr2022}     & 28.3\% & 7.8\% & 14.1\% & 14.9\% & 29.4\% & \textbf{55.0\%} & 64.46 M & 547.41 G & 92.1 ms \\
\midrule
\multirow{3}{*}{\rotatebox{90}{SOD}} 
& UFPMP-Det~\cite{ufpmpdet}             & 23.3\% & 5.9\% & 11.4\%& 8.9\% & 23.7\% & 31.7\% & 64.68 M  & 167.80 G  & 212.4 ms \\
& CEASC~\cite{Du_2023_CEASC}            & 17.2\% & 4.6\% & 6.1\% & 7.3\% & 20.1\% & 23.7\% & 43.05 M  & 103.21 G  & 31.5 ms \\
& CFINet~\cite{Yuan_2023_CFINet}        & 19.8\% & 5.4\% & 9.2\% & 8.0\% & 18.6\% & 27.7\% & 43.9 M   & 231.70 G  & 47.1 ms \\
\midrule
\multirow{7}{*}{\rotatebox{90}{YOLO}} 
& YOLOX-S~\cite{ge2021yolox}            & 17.3\% & 6.9\% & 8.2\% & 10.8\%& 19.5\% & 19.9\% & 8.9 M    & 34.1 G    & 24.3 ms \\
& YOLOv8-S~\cite{ultralytics} & 22.7\% & 9.0\% & 11.7\%& 11.6\%& 23.7\% & 36.0\% & 11.17 M  & 28.8 G    & 14.6 ms \\
& YOLOv8-L~\cite{ultralytics} & 26.7\% & 10.1\%& 14.4\%& 14.8\%& 26.2\% & 39.2\% & 43.69 M  & 165.7 G   & 26.4 ms \\
& YOLOv9-C~\cite{wang2402yolov9}        & 22.6\% & 7.7\% & 12.4\%& 10.9\%& 26.0\% & 39.2\% & 25.54 M  & 104.0 G   & 22.3 ms \\
& YOLO11-L~\cite{ultralytics}    & 21.1\% & 7.7\% & 12.3\%& 11.0\%& 26.0\% & 37.9\% & 25.30 M  & 87.6 G    & 18.2 ms \\
& HyperYOLO-S~\cite{HyperYOLO}          & 23.9\% & 9.2\% & 12.9\%& 10.8\%& 26.4\% & 39.4\% & 14.8 M   & 39.4 G    & 16.8 ms \\
& HyperYOLO-L~\cite{HyperYOLO}          & 26.2\% & 11.1\%& 14.5\%& 11.9\%& 28.3\% & 37.1\% & 56.2 M   & 212.0 G   & 31.2 ms \\
\midrule
\multirow{5}{*}{\rotatebox{90}{OVD}} 
& YOLOv8-S-worldv2\textsuperscript{ZS}~\cite{YOLOWorld}      & 10.2\% & 0.2\% & 1.2\% & 7.6\% & 18.3\% & 29.3\% & 12.7 M   & 33.5 G    & 11.5 ms \\
& YOLOv8-S-worldv2\textsuperscript{FT}~\cite{YOLOWorld}      & 22.0\% & 9.3\% & 12.3\%& 10.8\%& 26.0\% & 40.5\% & 12.7 M   & 33.5 G    & 11.5 ms \\
& YOLOv8-L-worldv2\textsuperscript{ZS}~\cite{YOLOWorld}      & 16.5\% & 1.5\% & 4.2\% & 12.2\%& 23.5\% & 36.9\% & 46.8 M   & 177.1 G   & 26.7 ms \\
& YOLOv8-L-worldv2\textsuperscript{FT}~\cite{YOLOWorld}      & 27.2\% & 11.0\%& 14.9\%& 14.4\%& 31.0\% & 50.4\% & 46.8 M   & 177.1 G   & 26.7 ms \\
& Grounding DINO~\cite{GroundingDINO}   & 21.7\% & 5.5\% & 8.0\% & 9.3\% & 22.4\% & 42.6\% & --       & --        & --      \\
\midrule
\multirow{2}{*}{\scriptsize Unified}
& Unicorn~\cite{Unicorn} & 27.6\% & 9.7\% & 12.8\% & 14.9\% & 28.8\% & 48.3\% & 58.18 M & 516.39 G & 67.8 ms \\
& UNINEXT~\cite{UNINEXT} & 25.4\% & 5.4\% & 11.4\% & 17.3\% & 32.7\% & 50.7\% & 195.86 M & 721.81 G & 198.2 ms \\
\midrule
\multirow{2}{*}{\scriptsize OURS}
& YOLOFT-S                              & 25.8\% & 9.8\% & 13.3\%& 14.4\%& 28.1\% & 39.2\% & 13.4 M   & 36.6 G    & 16.3 ms \\
& YOLOFT-L                              & \textbf{29.3\%} & \textbf{11.5\%}& \textbf{16.2\%}& \textbf{17.0\%}& \textbf{29.6\%}& 47.4\% & 49.8 M   & 194.5 G   & 29.3 ms \\
\addlinespace[3pt]
\bottomrule
\end{tabular}
\end{table*}

Object Detection aims to detect and localize multiple objects of predefined categories in a static image. For each image, the algorithm is expected to predict bounding boxes for all objects of the predefined categories, along with corresponding confidence scores. Video Object Detection (VOD) extends this task to video sequences, requiring algorithms to generate bounding boxes and confidence scores for each instance in every frame, if present. Unlike Object Detection (OD, denoted as DET), which focuses on individual images, VOD emphasizes temporal consistency across consecutive frames in video data. The same evaluation metrics as used in DET are applied to assess VOD performance. 
In the XS-VID benchmark, we conduct extensive evaluations and analyses of both DET and VOD methods to investigate effective methods for small-object detection.

\subsection{Data and Annotation}
In the detection tasks, the dataset contains 374 challenging video clips, including 308 for training (a total of 186,446 frames) and 66 for testing (a total of 36,478 frames). The number of objects per category is visualized in Fig.~\ref{fig:dataset-Statistics}(b) and   category imbalance presents a notable challenge to algorithm performance. For instance, in the training set, the number of car trajectories exceeds that of boats by more than six times.

\subsection{Review of Image and Video Object Detection Methods}
\subsubsection{General Object Detection Methods}
Object detection has evolved from anchor-based frameworks~\cite{Faster_RCNN, Cascade_RCNN, Cascade_RPN} to anchor-free paradigms. One-stage detectors~\cite{redmon2018yolov3, SSD} focus on real-time inference. Anchor-free methods like FCOS~\cite{tian2019fcos}, ATSS~\cite{zhang2020atss}, and YOLOX~\cite{ge2021yolox} eliminate anchors via keypoint regression or adaptive assignment. DyHead~\cite{dai2021dynamichead} unifies multi-scale contexts, while RepPoints~\cite{chen2020reppoints} reformulates detection as keypoint prediction. Transformer-based methods include Deformable-DETR~\cite{zhu2020deformabledetr} and Sparse R-CNN~\cite{sun2021sparsercnn}. Lightweight detectors include YOLOv9-C~\cite{wang2402yolov9}, YOLOv11~\cite{ultralytics}, Hyper-YOLO~\cite{HyperYOLO}, and D-FINE~\cite{DFINE2025}.

\subsubsection{Open-Vocabulary Object Detection}
Open-vocabulary detection (OVD)~\cite{OVD} detects unseen classes. Early works like ViLD~\cite{ViLD} use base-novel splits but are constrained by limited vocabularies. Vision-language pretraining~\cite{CLIP, ALIGN} motivates methods~\cite{RegionCLIP, Detic, DetPro, BARON, GLIP} that model OVD as image-text matching. OWL-ViTs~\cite{owlv1, owlv2} fine-tune ViTs for zero-shot detection, while Grounding DINO~\cite{GroundingDINO} and YOLO-World~\cite{YOLOWorld} extend grounding and real-time open-vocabulary detection, respectively.

\subsubsection{Video Object Detection}
FGFA~\cite{Zhu_2017_ICCVfgfa} enhances consistency through optical flow-guided aggregation. MEGA~\cite{Chen_2020_CVPRMEGA} utilizes long-term memory for global and local information. SELSA~\cite{wu2019selsa} aggregates features via attention-based fusion. TROI~\cite{TROI_2021} introduces temporal ROI alignment. DiffusionVID~\cite{diffusionvid} employs diffusion models for box refinement. StreamYOLO~\cite{yang2022realtimeStreamYOLO} improves real-time detection through dual-stream perception. TransVOD~\cite{9960850TransVOD} leverages DETR's query design, while YOLOV~\cite{shi2023yolov} focuses on key area selection.

\subsubsection{Small Object Detection}
Small-object methods commonly combine scale-aware augmentation~\cite{kisantal2019augmentation,yu2020scale}, feature super-resolution~\cite{9010998}, and cross-scale knowledge distillation~\cite{zhu2023scalekd}. Context-aware designs such as UFPMP-Det~\cite{ufpmpdet}, CEASC~\cite{Du_2023_CEASC}, and CFINet~\cite{Yuan_2023_CFINet} further improve fine-scale representation. Semantic Differentiation~\cite{SemDiff} and Unc-SOD~\cite{UncSOD} further address feature discrimination and uncertainty in small-object detection. TransVisDrone~\cite{sangam2023transvisdrone} uses spatiotemporal Transformers for drone detection.

\subsection{Evaluation Protocol}

We follow standard object detection metrics from MS COCO~\cite{COCO} and SODA~\cite{Cheng_2023SODA}, reporting Average Precision (AP) over 10 IoU thresholds from 0.50 to 0.95, with AP$_{50}$ and AP$_{75}$ denoting AP at IoU thresholds 0.50 and 0.75, respectively. The Detection Track evaluates category-labeled boxes for each frame and does not require target identities or trajectories; image-based and temporal detectors share this frame-level protocol. Small-object detection is assessed on area-based subsets: extra-small (AP$_{es}$, area $\le 12^2$ pixels), relatively-small (AP$_{rs}$, area $12^2$--$20^2$), general-small (AP$_{gs}$, area $20^2$--$32^2$), medium (AP$_m$, area $32^2$--$96^2$), and large (AP$_l$, area $> 96^2$). This notation concisely captures object size categories without a separate table.

\subsection{Results and Analysis}

Tab.~\ref{tab:xsvid_det_results} reports detector performance on XS-VID, exposing challenges in fine-grained spatial details and temporal consistency. Early VOD methods like DFF~\cite{Zhu_2017_CVPRdff} (10.9\% AP), FGFA~\cite{Zhu_2017_ICCVfgfa} (11.7\% AP), and SELSA~\cite{wu2019selsa} (13.9\% AP) achieve near-zero performance on extremely small objects ($\text{AP}_{es} = 0.0\%$). These methods sacrifice spatial details for high-level semantics through temporal aggregation, while incurring high computational cost (FGFA: 5579.10 G FLOPs). Recent VOD methods like StreamYOLO~\cite{yang2022realtimeStreamYOLO} (21.1\% AP) and TransVOD~\cite{9960850TransVOD} (17.4\% AP) improve efficiency but remain suboptimal.

General detectors excel on larger objects but struggle with small targets. ATSS~\cite{zhang2020atss}, RepPoints~\cite{chen2020reppoints}, and Sparse R-CNN~\cite{sun2021sparsercnn} achieve 23-24\% AP overall but only 6-8\% $\text{AP}_{es}$. D-FINE-L~\cite{DFINE2025} reaches 28.3\% AP through context modeling but requires 223.9 G FLOPs. SOD methods like CFINet~\cite{Yuan_2023_CFINet} lack temporal modeling, degrading performance under motion and occlusion.

OVD methods show clear generalization limits. YOLOv8-S-worldv2~\cite{YOLOWorld} achieves only 10.2\% AP zero-shot but reaches 22.0\% AP after fine-tuning. Grounding DINO~\cite{GroundingDINO} reaches 21.7\% AP but struggles with extremely small objects (5.5\% $\text{AP}_{es}$), highlighting the need for domain-specific adaptation.

YOLO detectors offer the best speed-accuracy trade-off. YOLOv8-L~\cite{ultralytics} achieves 26.7\% AP and 10.1\% $\text{AP}_{es}$ through multi-scale features, while HyperYOLO-L~\cite{HyperYOLO} reaches 26.2\% AP with improved efficiency. Building on these foundations, we propose YOLOFT, integrating Flow Mask Attention for feature alignment, 3D convolutions for temporal modeling, and SAM2-guided supervision. The detailed methodology and architecture of YOLOFT are comprehensively presented in Section~\ref{section:YOLOFT}.

Our proposed YOLOFT-L achieves 29.3\% AP (11.5\% $\text{AP}_{es}$, 16.2\% $\text{AP}_{rs}$, 17.0\% $\text{AP}_{gs}$) with 194.5 G FLOPs and 29.3 ms latency.

\subsection{Discussion}

XS-VID evaluation reveals two key limitations requiring future research.

\subsubsection{Temporal Modeling for Extremely Small Objects}
Our XS-VID evaluation reveals that traditional temporal methods (FGFA, SELSA) achieve $\text{AP}_{es} = 0.0\%$ due to spatial detail loss during feature fusion. We identify two key challenges: (1) \textbf{Scale-temporal trade-off}: Long-range temporal methods (MEGA, DiffusionVID) sacrifice fine spatial details critical for objects $<12^2$ pixels, and (2) \textbf{Motion alignment failure}: Small objects' high pixel displacement relative to their size makes optical flow-based alignment unreliable. Future work should explore scale-adaptive temporal modeling through multi-resolution temporal pyramids or scale-specific flow estimation.

\subsubsection{Dense Feature Learning for Small-Object Videos}
XS-VID's high density of extremely small objects (29.4\% $<12^2$ pixels) exposes standard detection losses' inadequacy. Our analysis shows auxiliary supervision provides +2.2\% $\text{AP}_s$ improvement, indicating that dense pixel-level guidance is crucial for small-object representation learning. Future work may investigate pretraining schemes that strengthen fine-grained spatial representations without requiring additional annotations.

\section{SOT TRACK}

The Single Object Tracking (SOT) track focuses on locating a specific object throughout a video sequence, given its bounding box in the first frame.

\subsection{Data and Annotation}

\begin{figure}[t]
    \centering
    \includegraphics[width=1.0\linewidth]{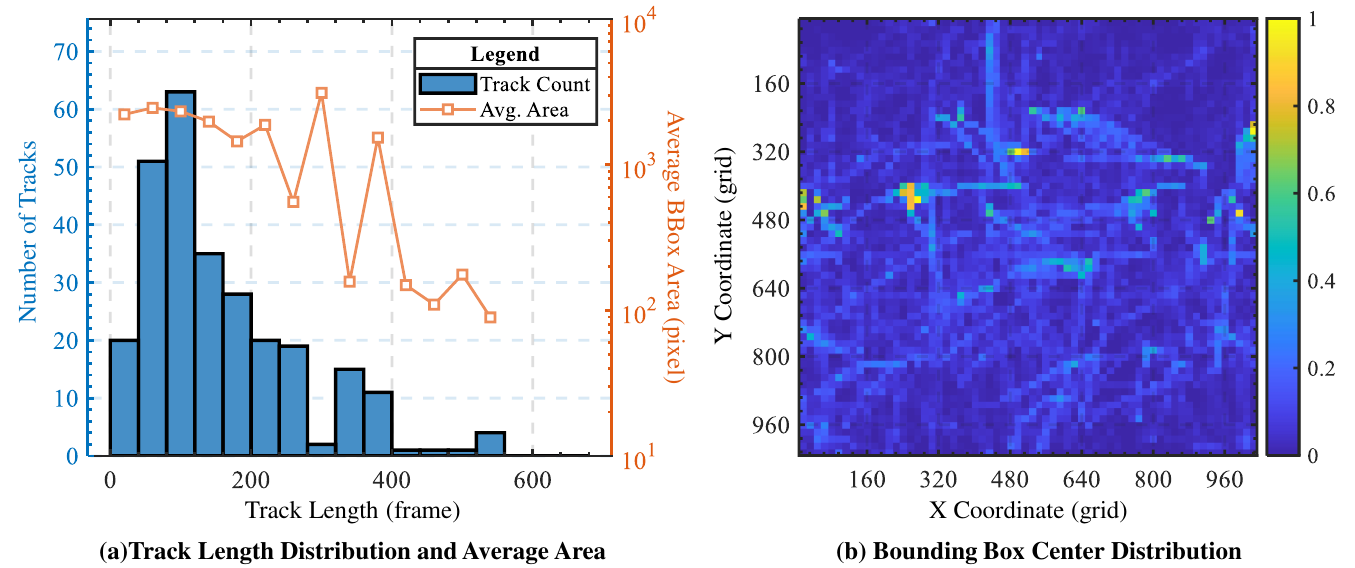}
    \caption{
    \textbf{Statistical analysis of single object tracking (SOT) trajectories in our dataset.}
    (a) Histogram of track lengths and the corresponding average bounding box areas. The bars represent the number of trajectories within each length interval, while the red curve indicates the mean object size in each bin. 
    (b) Spatial distribution heatmap of object centers aggregated across all frames. The relatively uniform coverage across the image space reflects diverse scene compositions and helps mitigate location bias.
    }
    \label{fig:sot_fig}
\end{figure}

We use the same training set as the Detection Track, consisting of 308 video clips. Participants can freely choose how to train their models on this data. For testing, we do not include all trajectories from the 66 test videos. Since online SOT methods typically focus on frame-by-frame association for a single target, we select up to five of the longest eligible trajectories per video and exclude those dominated by prolonged full occlusion. This selection yields \textbf{272} long-trajectory single-object tracking sequences, covering a total of 107{,}788 frames. Further split and evaluation details are provided in the SM's Sec.~II-C. Fig.~\ref{fig:sot_fig} visualizes key properties of the test data.

\subsection{Review of Single Object Tracking Methods}

Tracking methods fall into image-level trackers using appearance features and video-level trackers exploiting temporal context. Classical trackers~\cite{SINT, SiameseFC, SiameseRPN, SiamRPNplusplus, Deeper-wider-SiamRPN, SiamFC++, SiamRCNN, AiATrack, transinmo} initialize with first-frame target appearance and locate objects via similarity matching. They degrade under deformation, motion blur, or illumination changes. Methods~\cite{stmtrack, TMT, Stark, keeptrack, seqtrack} incorporate online-updated templates to improve robustness but remain appearance-constrained.

Video-level frameworks better leverage temporal information. TCTrack~\cite{tcttrack} integrates temporal context via adaptive convolutions. VideoTrack~\cite{videotrack} introduces video transformer backbones. HIPTrack~\cite{HIPTrack} injects historical foreground masks as prompts. PVT++~\cite{PVT++} fuses visual and motion features for delay-resilient inference. LoRAT~\cite{lorat} uses parameter-efficient tuning for large ViT backbones.

Another line of work focuses on token-based spatiotemporal modeling. ODTrack~\cite{odtrack} propagates token sequences to associate frame-wise object context. ARTrack~\cite{artrack} and ARTrackV2~\cite{artrackv2} encode historical coordinates as auxiliary tokens. SeqTrack~\cite{seqtrack} leverages generative decoding over trajectory-token fusion. AQATrack~\cite{aqatrack} uses learnable autoregressive queries to mine temporal dependencies, while EVPTrack~\cite{evptrack} applies spatio-temporal markers for frame-to-frame propagation. MCITrack~\cite{mcitrack} enhances contextual representation by capturing richer and more discriminative temporal information, leading to improved tracking accuracy and robustness.

\subsection{Evaluation Protocol}

We evaluate trackers following widely adopted protocols, including OTB-2015~\cite{ObjectTrackingBenchmarkWu2015} and LaSOT~\cite{Lasot}, using a one-pass evaluation (OPE) on all sequences. Two standard metrics are used: precision and success. Precision measures the center location error between predicted and ground-truth bounding boxes, reported as the percentage of frames with an error within a 20-pixel threshold. Success is measured by the intersection over union (IoU) between predicted and ground-truth boxes, with the success curve plotted over IoU thresholds from 0 to 1; the area under this curve (AUC) is used to rank trackers. In addition, we report each tracker's frames per second (FPS) to assess its practical efficiency for real-time applications.

\begin{table}[!ht]
    \centering
    \caption{Comparison on the XS-VID SOT track. AUC is the area under the success curve, and precision is measured at the 20-pixel threshold. Specialized trackers and unified frameworks are grouped separately.}
    \label{tab:sot_methods_comparison}
    \scriptsize
    \setlength{\tabcolsep}{3pt}
    \resizebox{\columnwidth}{!}{%
    \begin{tabular}{lrrrrr}
        \toprule
        Tracker & AUC (\%) & Precision (\%) & Speed (FPS) & Pub. & Year \\
        \midrule
        \multicolumn{6}{c}{\textit{Specialized trackers}} \\
        \midrule
        SeqTrack\textsubscript{b256}\cite{seqtrack} & 52.94 & 76.96 & 36 & CVPR & 2023 \\
        SeqTrack\textsubscript{l384}\cite{seqtrack} & 54.26 & 78.34 & 7.6 & CVPR & 2023 \\
        PVT++\cite{PVT++} & 11.49 & 25.71 & 36.2 & ICCV & 2023 \\
        LoRAT\textsubscript{b224}\cite{lorat} & 62.41 & 83.31 & \textbf{209} & ECCV & 2024 \\
        LoRAT\textsubscript{g224}\cite{lorat} & \textbf{66.40} & 87.07 & 50 & ECCV & 2024 \\
        ODTrack\textsubscript{base}\cite{odtrack} & 58.50 & 85.30 & 28.55 & AAAI & 2024 \\
        ODTrack\textsubscript{large}\cite{odtrack} & 61.25 & \textbf{89.11} & 15.67 & AAAI & 2024 \\
        HIPTrack\cite{HIPTrack} & 57.75 & 81.93 & 46.5 & CVPR & 2024 \\
        MCITrack\textsubscript{s224}\cite{mcitrack} & 53.63 & 76.17 & 19.8 & AAAI & 2025 \\
        MCITrack\textsubscript{l384}\cite{mcitrack} & 56.53 & 79.88 & 8.5 & AAAI & 2025 \\
        SiamFC~\cite{SiameseFC} & 42.42 & 69.85 & 118.0 & ECCVW & 2016 \\
        MDNet~\cite{MDNet} & 45.59 & 72.83 & 2.7 & CVPR & 2016 \\
        ToMP50~\cite{ToMP} & 53.51 & 78.42 & 17.0 & CVPR & 2022 \\
        TransT~\cite{TransT} & 57.31 & 80.72 & 22.0 & CVPR & 2021 \\
        \midrule
        \multicolumn{6}{c}{\textit{Unified frameworks}} \\
        \midrule
        Unicorn~\cite{Unicorn} & 25.62 & 45.83 & 5.1 & ECCV & 2022 \\
        UNINEXT~\cite{UNINEXT} & 38.84 & 54.30 & 21.0 & CVPR & 2023 \\
        YOLOFT-L & \textbf{59.63} & \textbf{81.35} & 17.1 & -- & -- \\
        \bottomrule
    \end{tabular}
    }
\end{table}

\subsection{Results and Analysis}

We evaluate recent single object tracking (SOT) algorithms on the XS-VID benchmark, and Table~\ref{tab:sot_methods_comparison} summarizes their performance. Among these methods, video-level Transformer-based trackers achieve the best results. LoRAT\textsubscript{g224}~\cite{lorat} attains the highest area under the curve (AUC) of \textbf{66.4\%} and precision of \textbf{87.07\%}, highlighting the strong capability of large ViT backbones in small-object tracking. Its lightweight variant, LoRAT\textsubscript{b224}~\cite{lorat}, reaches \textbf{209 FPS}, demonstrating suitability for real-time deployment. Trackers leveraging temporal context, such as ODTrack~\cite{odtrack} and HIPTrack~\cite{HIPTrack}, also show competitive performance. In particular, ODTrack\textsubscript{large} achieves the highest precision (\textbf{89.11\%}), indicating that token sequence propagation effectively maintains tracking robustness under occlusion and appearance changes.

Recent methods incorporating memory-based spatiotemporal modeling and autoregressive decoding, such as MCITrack~\cite{mcitrack} and SeqTrack~\cite{seqtrack}, maintain a balanced trade-off between accuracy and efficiency. In contrast, image-level trackers like PVT++~\cite{PVT++}, while computationally efficient, exhibit limitations in complex small-object scenes, highlighting the importance of temporal modeling beyond appearance-only matching.

The expanded comparison also includes SiamFC~\cite{SiameseFC}, MDNet~\cite{MDNet}, ToMP50~\cite{ToMP}, and TransT~\cite{TransT}. Following the unified designs of Unicorn~\cite{Unicorn} and UNINEXT~\cite{UNINEXT}, we use YOLOFT-L for unified extension verification; it obtains 59.63\% AUC on XS-VID.

\subsection{Discussion}

XS-VID exposes fundamental limitations in small-object tracking, where conventional appearance-based methods fail due to insufficient visual features. Below, we highlight two key future research directions.

\subsubsection{Long-Term Association for Small Objects}
Our evaluation reveals that image-level trackers (PVT++) achieve only 25.71\% precision compared to video-level trackers (LoRAT: 87.07\%) on small objects. Template updates designed for medium-large objects become ineffective when resolution drops below 20×20 pixels, and appearance-only matching cannot sustain association across hundreds of frames. Future work should explore scale-adaptive template fusion and motion-predictive association strategies~\cite{TMT, artrack}.

\subsubsection{Scale-Aware Tracking Architectures}
XS-VID shows that temporal context becomes essential for small objects where spatial features are insufficient, with video-level trackers outperforming image-level methods by 40+ percentage points. Future work could study self-supervised pretraining on small-object sequences and multi-scale temporal templates for sustained small-object tracking.

\section{MOT TRACK}

The multi-object tracking (MOT) task aims to recover the trajectories of multiple objects throughout a video sequence. We adopt the general MOT setting, also known as class-agnostic tracking~\cite{dave2020tao}. This track utilizes the same dataset as the Detection Track, comprising 308 video clips for training and 66 for testing. The evaluation focuses on six object categories: \emph{Person}, \emph{Car}, \emph{Bicycle}, \emph{Cyclist}, \emph{Boat}, and \emph{Truck}, while ignoring all other classes and any regions marked with the \texttt{ignore} label.

\subsection{Data and Annotation}

\begin{figure}[ht]
    \centering
    \includegraphics[width=1.0\linewidth]{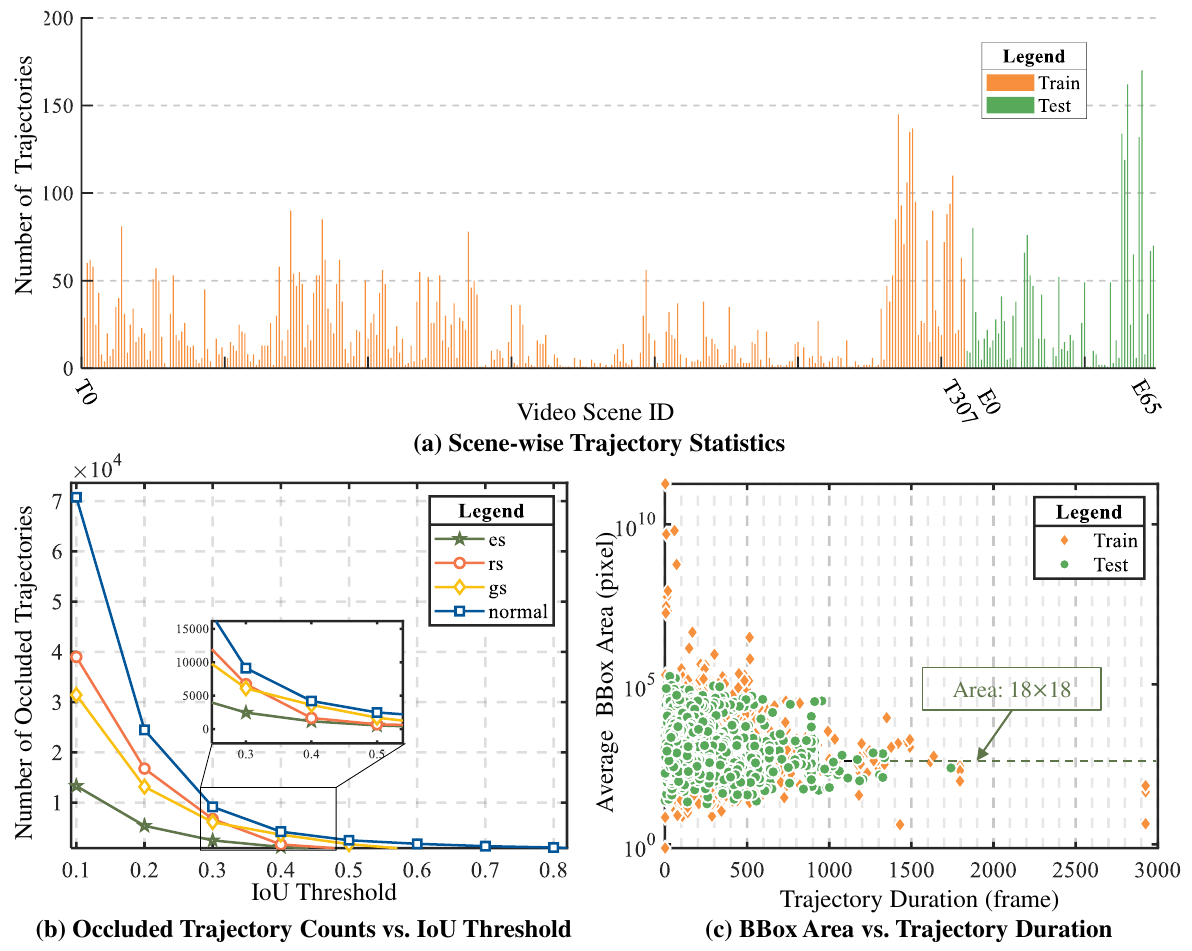}
    \caption{
    \textbf{Trajectory-level statistics of the XS-VID MOT Track.}
    (a) Scene-wise trajectory counts across 374 videos (T0--T307 training, E0--E65 testing) show variations in object density.
    (b) Occluded trajectories under different IoU thresholds. When \textbf{IoU < 0.3}, \textbf{RS-scale trajectories} are more frequently occluded than GS or ES, increasing identity switch likelihood.
    (c) Trajectory duration vs. average bounding box area. Most trajectories cluster around \textbf{18$\times$18 pixels}, with \textbf{long-duration trajectories (>250 frames)} remaining in the small-object regime, highlighting the challenge of long-term association for tiny objects.
    }
    \label{fig:MOT_track}
\end{figure}

The XS-VID MOT Track follows the TAO annotation format~\cite{dave2020tao}, with object-level annotations grouped into trajectories. As shown in Fig.~\ref{fig:MOT_track}, trajectory durations range from 1 to 2{,}926 frames; 15.34\% are shorter than 25 frames, while 17.95\% exceed 250 frames. Many long-duration trajectories remain in the small-object regime, and the high occlusion rate of \textit{rs} trajectories further increases the risk of identity switches. These properties emphasize long-term association under small-object constraints.

\subsection{Review of Multi-Object Tracking Methods}

Track-by-detection has become the prevailing paradigm in multi-object tracking (MOT), where objects are first detected in individual frames and then temporally associated to form trajectories.

ByteTrack~\cite{ByteTrack} represents a significant milestone by introducing low-score detections and a two-stage association strategy, substantially improving robustness against occlusion and missed detections. OC-SORT~\cite{ocsort} builds upon this foundation by incorporating Kalman filter-based motion modeling and local trajectory refinement, enhancing short-term motion estimation.

DeepOCSORT~\cite{deepocsort} further strengthens the framework by integrating deep ReID features and occlusion-aware modules, improving discrimination in crowded scenes. StrongSORT~\cite{Strongsort}, as an upgraded version of OC-SORT, adopts feature fusion, occlusion modeling, and enhanced appearance encoding to increase association robustness.

In parallel, BoostTrack~\cite{BoostTrack} fuses multi-source cues from appearance and motion via a temporal boosting mechanism, delivering strong performance across various benchmarks. BoT-SORT~\cite{botsort} leverages a strong ReID backbone from BoT, combined with Gaussian-based Hungarian matching and trajectory confidence modeling, to achieve high-accuracy tracking under complex scenarios.

The expanded evaluation additionally covers GNN association with SUSHI~\cite{SUSHI}, end-to-end Transformer trackers including TrackFormer~\cite{TrackFormer}, MOTR~\cite{MOTR}, and MOTIP~\cite{MOTIP}, and the unified frameworks Unicorn~\cite{Unicorn} and UNINEXT~\cite{UNINEXT}.

\subsection{Evaluation Protocol}

We adopt the evaluation protocol proposed in the TAO benchmark~\cite{dave2020tao}, using the mean Average Precision (mAP) computed with 3D Intersection over Union (IoU) to assess multi-object tracking performance in the wild. This metric offers a concise and interpretable measure of tracking quality, particularly suitable for large-scale multi-category scenarios, as demonstrated by its usage in recent tracking datasets~\cite{deng2009imagenet,yang2019video}.

Furthermore, prior analysis~\cite{dave2020tao} shows that 3D IoU correlates strongly with identity consistency, a core aspect of multi-object tracking. Specifically, it aligns closely with identity-related errors such as identity switches, which are quantified by standard MOT challenge metrics~\cite{milan2016mot16}. This reinforces the effectiveness of 3D IoU as a unified criterion for evaluating both spatial and identity accuracy in multi-object tracking systems.

Main Tab.~\ref{tab:mot_results} reports TAO mAP, MOTA, and IDF1. The complete metric breakdown and evaluation settings are provided in the SM's Sec.~II-D.

\begin{figure*}[ht]

    \centering
    \includegraphics[width=1.0\linewidth]{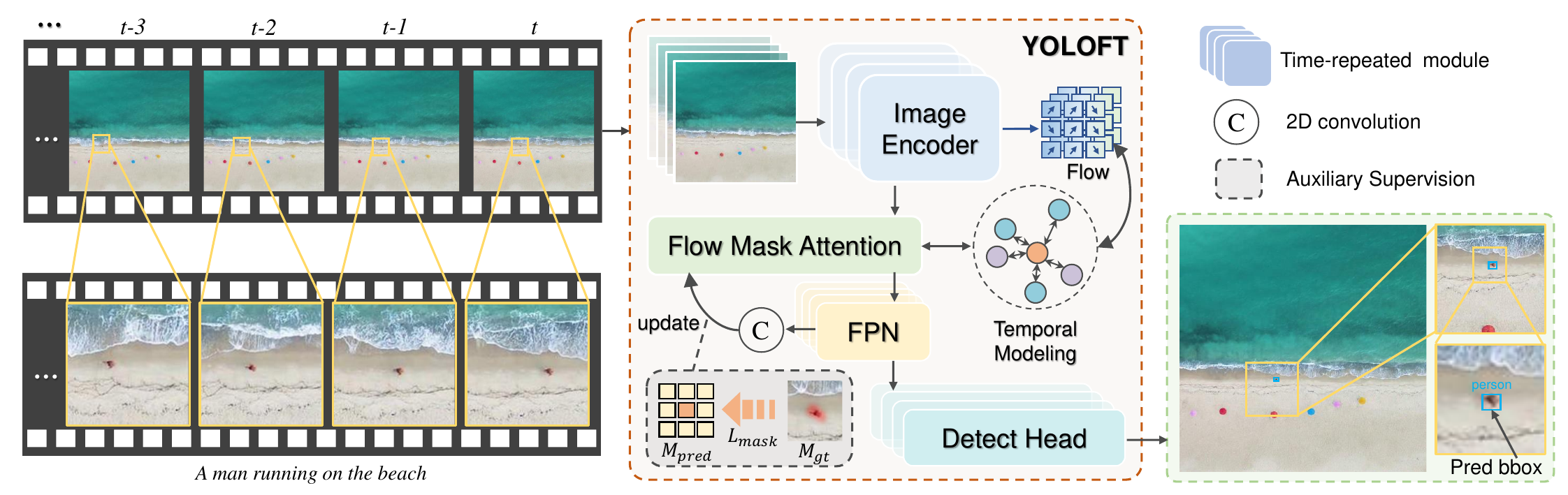}
    \caption{\textbf{Overview of the proposed YOLOFT architecture.} The framework consists of three key components: (1) a lightweight 3D convolutional module for efficient local temporal modeling across adjacent video frames; (2) a Flow Mask Attention (FMA) module that dynamically predicts flow-guided offsets, attention weights, and fusion gates to improve temporal alignment and feature fusion; and (3) a mask-assisted auxiliary supervision branch that leverages motion-aware binary masks to provide targeted spatial supervision, enabling more efficient optimization and better localization of small objects.}

    \label{fig:FrameworkYOLOFT}
\end{figure*}

\subsection{Results and Analysis}

We expand the MOT benchmark across tracking-by-detection, GNN association, end-to-end Transformer tracking, and unified frameworks in Tab.~\ref{tab:mot_results}. BoT-SORT with YOLOFT-L detections reaches the highest TAO mAP of 32.1\%. Following Unicorn and UNINEXT, we use YOLOFT-L for unified extension verification; it obtains 31.5\% TAO mAP and 55.2\% IDF1 on MOT, together with 59.63\% AUC on SOT. The complete MOT results and unified-evaluation settings are provided in the SM's Secs.~II-D--II-E.

\subsection{Discussion}

XS-VID highlights persistent challenges in multi-object tracking of small objects. We identify two key directions:

\subsubsection{Scale-Dependent Association}
The expanded benchmark shows substantial performance differences across association paradigms on extremely small objects. Current ReID features trained on larger objects fail for sub-32$^2$ targets, causing frequent identity switches. Future work should focus on motion-centric association~\cite{ocsort} and scale-aware ReID.

\subsubsection{Small-Object Identity Preservation}
High object density exposes limitations of appearance-based matching. BoT-SORT with YOLOFT-L detections reaches 32.1 TAO mAP, while unified YOLOFT-L obtains 55.2 IDF1. Joint detector-tracker optimization~\cite{TrackFormer} and specialized small-object metrics may alleviate these issues.

\begin{table}[!t]
    \centering
    \renewcommand{\arraystretch}{1.1}
    \caption{Expanded MOT evaluation on XS-VID. T-by-D, GNN, and E2E-Trans denote tracking by detection, graph neural network association, and end-to-end Transformer tracking, respectively. Detailed results and evaluation settings are provided in the SM's Sec.~II-D.}
    \label{tab:mot_results}
    \footnotesize
    \setlength{\tabcolsep}{2.8pt}
    \resizebox{0.95\columnwidth}{!}{%
    \begin{tabular}{@{}l l r r r@{}}
        \toprule
        Method & Paradigm & TAO mAP & MOTA & IDF1 \\
        & & (\%) & (\%) & (\%) \\
        \midrule
        \multicolumn{5}{c}{\textit{Multi-stage with YOLOFT detections}} \\
        \midrule
        BoostTrack~\cite{BoostTrack} & T-by-D & 14.4 & 17.3 & 40.3 \\
        SUSHI~\cite{SUSHI} & GNN & 15.1 & 26.7 & 40.5 \\
        HybridSORT~\cite{HybridSORT} & T-by-D & 16.5 & \textbf{37.5} & 48.7 \\
        OC-SORT~\cite{ocsort} & T-by-D & 20.1 & 34.8 & 47.8 \\
        ByteTrack~\cite{ByteTrack} & T-by-D & 22.3 & 36.1 & 52.6 \\
        StrongSORT~\cite{Strongsort} & T-by-D & 26.0 & 36.8 & 53.9 \\
        DeepOCSORT~\cite{deepocsort} & T-by-D & 26.9 & 37.4 & 53.0 \\
        BoT-SORT~\cite{botsort} & T-by-D & \textbf{32.1} & 37.1 & \textbf{54.7} \\
        \midrule
        \multicolumn{5}{c}{\textit{End-to-end and unified}} \\
        \midrule
        MOTIP~\cite{MOTIP} & E2E-Trans & 6.5 & 13.2 & 36.2 \\
        MOTR~\cite{MOTR} & E2E-Trans & 13.0 & 19.8 & 35.7 \\
        TrackFormer~\cite{TrackFormer} & E2E-Trans & 15.2 & 27.6 & 41.2 \\
        UNINEXT~\cite{UNINEXT} & Unified & 20.1 & \textbf{32.6} & 50.2 \\
        Unicorn~\cite{Unicorn} & Unified & 24.6 & 32.4 & 51.2 \\
        YOLOFT & Unified & \textbf{31.5} & 30.7 & \textbf{55.2} \\
        \bottomrule
    \end{tabular}
    }
\end{table}

\section{YOLOFT}\label{section:YOLOFT}

The overall architecture of YOLOFT is illustrated in Fig.~\ref{fig:FrameworkYOLOFT}. Built upon the YOLOv8 detector~\cite{ultralytics}, YOLOFT is designed to address the challenges of small-object detection in video detection scenes. It comprises two main components: a \textbf{3D Convolutional Temporal Modeling} module for local motion capture between adjacent frames, and the proposed \textbf{Flow Mask Attention (FMA)} module that enhances feature consistency across frames. The XS-VID analysis motivates complementary design choices: the design study in Sec.~VIII-B examines convolution, upsampling, and localization choices for fine-scale representation, while short-term 3D convolution, the proposed FMA mechanism, and mask-assisted supervision respectively provide local temporal context, motion-aligned aggregation, and denser object-region guidance. Their roles are examined through the design studies and ablations in Secs.~VIII-B--VIII-C.

\paragraph{Local Temporal Encoder} 
To capture short-term motion cues, YOLOFT employs a 3D convolutional block~\cite{I3D, X3D} applied to the previous temporal feature $\hat{F}_{t-1}$ and the current frame feature $F_t$. Unlike long-range temporal modeling in video recognition~\cite{SlowFast, Girdhar_2019_CVPR}, this local design mitigates error accumulation and reduces computational overhead. Specifically,
\begin{equation}
\hat{F}_t = \text{3DConv}(\text{Stack}(\hat{F}_{t-1}, F_t)),
\end{equation}
where $\hat{F}_t$ encodes spatial semantics and localized temporal dynamics. Here, $\hat{F}_{t-1}$ is not a raw feature from frame $t-1$, but the temporally encoded representation obtained in the previous step, thus providing memory that accumulates temporal context while remaining lightweight.

\paragraph{Dynamic Temporal Feature Fusion}
The Flow Mask Attention (FMA) module is inspired by optical flow~\cite{teed2020raft} and dynamic convolution~\cite{DCNv3, DCNv4, fdconv}. Its goal is to align historical and current features to handle motion, blur, and occlusion. Given the temporal memory $\hat{F}_t$ and the current feature $F_t$, FMA predicts three dynamic components:
\begin{itemize}
  \item \textbf{Flow Offsets} $\Delta p_{g,k}$: pixel-level displacements estimated from $\hat{F}_t$ via depthwise convolution;
  \item \textbf{Attention Weights} $M_{g,k}$: normalized weights derived from $\hat{F}_t$, enabling spatially adaptive aggregation;
  \item \textbf{Fusion Gate} $W$: a learnable parameter that adaptively balances the contributions of $F_t$ and the motion-aligned historical signal $H_t$.
\end{itemize}

The motion-aligned historical feature is then computed as:
\begin{equation}
H_t(p) = \sum_{g=1}^{G} \sum_{k=1}^{K^2} M_{g,k}(p) \cdot F_t(p + p_k + \Delta p_{g,k}(p)),
\end{equation}
where both $M$ and $\Delta p$ are generated from $\hat{F}_t$, ensuring that historical context guides the alignment of the current-frame feature map. Finally, the fused representation is:
\begin{equation}
F'_t(p) = F_t(p) \odot (1 - W(p)) + H_t(p) \odot W(p),
\end{equation}
which adaptively integrates motion-aligned signals with the current frame, leading to robust detection under temporal variations.

\begin{figure}[ht]
    \centering
    \includegraphics[width=1.0\linewidth]{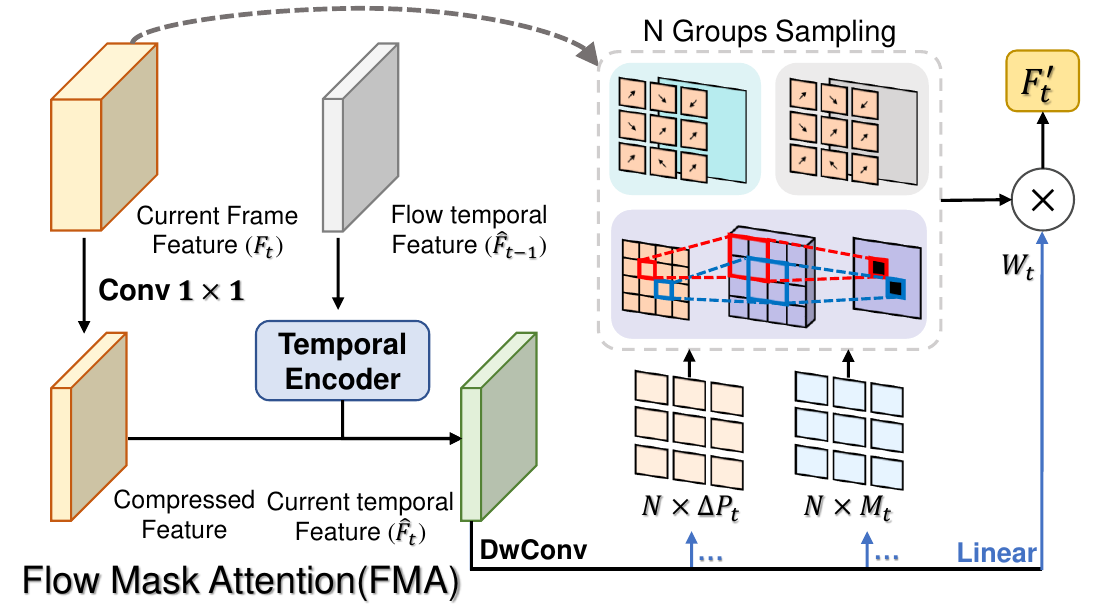}
    \caption{\textbf{The Flow Mask Attention (FMA) module combines temporal modeling with spatial feature refinement to adaptively enhance the current frame feature $\mathbf{F_t}$.} It uses a temporal encoder to capture short-term motion between $F_t$ and the previous temporal feature $\hat{F}_{t-1}$, producing an updated temporal feature $\hat{F}_t$.} 
    \label{fig:FMA}
\end{figure}

\begin{table*}[!t]
\centering
\caption{Performance comparison on VisDroneVID validation set. All AP values are reported in \%. FLOPs measured at input size $640^2$.}
\label{tab:visdrone_det_results}
\begin{tabular}{llrrrrrrrrr}
\toprule
\multicolumn{2}{c}{\textbf{Method}} & $\text{AP}^{val}$ & $\text{AP}^{val}_{es}$ & $\text{AP}^{val}_{rs}$ & $\text{AP}^{val}_{gs}$ & $\text{AP}^{val}_{m}$ & $\text{AP}^{val}_{l}$ & \#Params. & FLOPs & Latency$\scriptstyle [bs=1]$ \\
\midrule
\multirow{6}{*}{\rotatebox{90}{VOD}} 
& DFF~\cite{Zhu_2017_CVPRdff}             & 10.3\% & 0.0\% & 0.1\% & 3.4\% & 13.6\% & 21.8\% & 120.0 M & 458.7 G & 25.5 ms \\
& FGFA~\cite{Zhu_2017_ICCVfgfa}           & 13.6\% & 0.0\% & 0.9\% & 6.3\% & 17.8\% & 28.5\% & 122.4 M & 7874.6 G & 181.8 ms \\
& SELSA~\cite{wu2019selsa}          & 11.8\% & 0.0\% & 0.5\% & 2.7\% & 14.3\% & 30.2\% & 128.4 M & 6917.7 G & 110.0 ms \\
& TROI~\cite{TROI_2021}            & 12.0\% & 0.0\% & 0.1\% & 4.8\% & 16.6\% & 24.7\% & 136.0 M & 7487.7 G & 285.7 ms \\
& TransVOD~\cite{9960850TransVOD}       & 9.7\%  & 1.0\% & 3.2\% & 4.9\% & 11.5\% & 23.8\% & 33.8 M  & 613.9 G & 136.0 ms \\
& StreamYOLO~\cite{yang2022realtimeStreamYOLO}     & \textbf{18.0\%} & 1.6\% & 5.1\% & 10.6\% & \textbf{22.3\%} & 33.9\% & 54.8 M  & 698.1 G & 47.5 ms \\
\midrule
\multirow{7}{*}{\rotatebox{90}{GOD}} 
& FCOS~\cite{tian2019fcos}           & 12.4\% & 1.3\% & 3.1\% & 4.8\% & 13.8\% & 30.6\% & 32.1 M  & 202.0 G & 31.8 ms \\
& ATSS~\cite{zhang2020atss}           & 13.7\% & 1.5\% & 4.6\% & 7.2\% & 16.2\% & 29.9\% & 32.1 M  & 206.0 G & 34.9 ms \\
& DyHead~\cite{dai2021dynamichead}         & 9.3\%  & 1.4\% & 3.5\% & 5.0\% & 10.7\% & 20.7\% & 38.9 M  & 112.0 G & 98.0 ms \\
& RepPoints~\cite{chen2020reppoints}      & 13.6\% & 0.7\% & 3.9\% & 5.4\% & 16.3\% & 29.0\% & 36.8 M  & 194.0 G & 37.8 ms \\
& Deformable-DETR~\cite{zhu2020deformabledetr}& 9.8\%  & 2.5\% & 3.7\% & 5.1\% & 11.9\% & 19.5\% & 40.1 M  & 200.0 G & 52.3 ms \\
& Sparse RCNN~\cite{sun2021sparsercnn}    & 8.1\%  & 1.0\% & 2.9\% & 4.5\% & 9.5\%  & 16.0\% & 106.0 M & 181.3 G & 41.8 ms \\
& Cascade RCNN~\cite{Cascade_RCNN}    & 12.5\% & 0.9\% & 3.9\% & 6.2\% & 15.1\% & 25.3\% & 41.4 M  & 196.7 G & 45.3 ms \\
\midrule
\multirow{2}{*}{\rotatebox{90}{SOD}} 
& CEASC~\cite{Du_2023_CEASC}            & 11.0\% & 1.7\% & 3.5\% & 4.4\% & 13.0\% & 23.8\% & 43.1 M  & 103.2 G & 31.0 ms \\
& CFINet~\cite{Yuan_2023_CFINet}         & 12.2\% & 1.0\% & 3.3\% & 6.3\% & 15.1\% & 25.8\% & 43.9 M  & 231.7 G & 47.1 ms \\
\midrule
\multirow{4}{*}{\rotatebox{90}{YOLO}} 
& YOLOX-S~\cite{ge2021yolox}         & 7.8\%  & 1.6\% & 3.5\% & 5.6\% & 10.4\% & 12.8\% & 8.9 M   & 34.1 G  & 24.0 ms \\
& YOLOv8-S~\cite{ultralytics}       & 13.2\% & 3.9\% & 5.0\% & 10.1\%& 16.1\% & 22.9\% & 11.17 M  & 28.8 G    & 14.6 ms \\
& YOLOv8-L~\cite{ultralytics}       & 16.0\% & 3.6\% & 5.1\% & 9.9\% & 19.7\% & 27.3\% & 43.69 M  & 165.7 G   & 26.4 ms \\
& YOLOv9-C~\cite{wang2402yolov9}       & 15.5\% & 1.8\% & 5.8\% & 9.8\% & 19.1\% & 33.4\% & 25.54 M  & 104.0 G   & 22.3 ms \\
\midrule
\multirow{2}{*}{\rotatebox{90}{OURS}} 
& YOLOFT-S       & 16.0\% & 2.4\% & 5.4\% & 9.6\%&  18.3\% & 28.4\% & 13.4 M  & 36.3 G  & 16.3 ms \\
& YOLOFT-L       & 17.8\% & \textbf{3.9\%} & \textbf{6.0\%} & \textbf{11.2\%} & 20.5\% & \textbf{34.6\%} & 49.8 M  & 194.5 G & 29.3 ms \\
\addlinespace[3pt]
\bottomrule
\end{tabular}
\end{table*}

\paragraph{Mask-Assisted Auxiliary Supervision} 
Standard detection losses, including $L_{cls}$, $L_{box}$, and $L_{dfl}$, often yield sparse temporal gradients—particularly for extremely small or occluded objects. To address this, we introduce a mask-assisted auxiliary supervision branch that provides dense, pixel-level guidance for temporal feature learning.

We employ pseudo masks generated by SAM2~\cite{ravi2024sam2} on the training set (e.g., XS-VID) to indicate object regions. A lightweight segmentation head takes the fused feature $F'_t$ as input and predicts a coarse binary mask. The mask loss $L_{mask}$ is defined as a pixel-wise cross-entropy loss, computed only within matched ground-truth bounding boxes:
\begin{equation}
L_{\text{total}} = L_{\text{det}} + \lambda L_{\text{mask}}.
\end{equation}
This auxiliary objective encourages the model to focus on motion-consistent object regions, facilitating more stable optimization and improved localization of small targets.

\paragraph{Unified Model Extension}
Drawing on the staged training paradigm of Unicorn~\cite{Unicorn} and UNINEXT~\cite{UNINEXT}, we extend YOLOFT for evaluation on the Detection, SOT, and MOT tracks. The corresponding setting is provided in the SM's Sec.~II-E.

\vspace{0.5em}
\noindent
YOLOFT presents a compact and efficient architecture tailored for small-object detection in videos. It leverages short-term 3D convolutions for local temporal modeling, FMA for motion-aligned dynamic fusion, and mask-assisted supervision to compensate for weak training signals. Together, these components enable robust and consistent detection performance across diverse and long-duration video scenarios beyond traditional aerial or traffic-focused domains.

\section{Experiments}\label{section:Experiments}

We present three experimental studies: YOLOFT evaluation on XS-VID and VisDrone (Sec.~\ref{sec:e1}), design considerations for small-object video detection (Sec.~\ref{sec:e2}), and YOLOFT's ablation studies (Sec.~\ref{sec:e3}).

\subsection{Comparison with State-of-the-art Detection Methods}
\label{sec:e1}

In addition to the Detection Track section (Sec.~\ref{sec:VOD_TRACK} and Tab.~\ref{tab:xsvid_det_results}), we evaluate the generalization performance of state-of-the-art models on the VisDrone-VID dataset. Tab.~\ref{tab:visdrone_det_results} summarizes the results. Since we already report AP$_{es}$, AP$_{rs}$, and AP$_{gs}$, the coarse AP$_s$ metric is omitted. We exclude TransVisDrone~\cite{sangam2023transvisdrone} due to its high inference latency ($>$2 s/frame), which makes it unsuitable for real-time applications.

\subsubsection{XS-VID Presents a High-Resolution Challenge for Fine-Grained Detection}

As shown in Tab.~\ref{tab:dataset_detail}, XS-VID includes far more extremely small (29.4\%) and relatively small (26.1\%) objects than VisDrone (0.5\% and 7.5\%). In ImageNetVID, these categories are nearly absent~\cite{russakovsky2015imagevid}. This distribution narrows performance differences on VisDrone, where medium-sized objects dominate. In contrast, XS-VID requires fine spatial resolution, temporal consistency, and robustness to occlusion and low contrast, which reveals the limitations of conventional detectors like DFF~\cite{Zhu_2017_CVPRdff} and FGFA~\cite{Zhu_2017_ICCVfgfa}. These models struggle to capture fine-scale features, resulting in AP$_{es}$ scores near zero. This highlights the need for architectures specifically designed for small objects, such as YOLOFT.

\subsubsection{YOLOFT Bridges the Gap Between Temporal Modeling and Efficiency}

YOLOFT mitigates key limitations of previous detectors by incorporating three efficient components: Flow Mask Attention (FMA), short-term 3D temporal modeling, and mask-guided auxiliary supervision. Unlike prior VOD methods that rely on long-range temporal fusion and incur high computational cost (e.g., FGFA~\cite{Zhu_2017_ICCVfgfa}, SELSA~\cite{wu2019selsa}), YOLOFT-L achieves 29.3\,ms latency while achieving strong performance on both XS-VID and VisDrone. As shown in Fig.~\ref{fig:bubble_model}, both YOLOFT-S and YOLOFT-L achieve a favorable trade-off between accuracy and speed. This performance is driven by motion-guided attention and mask-based supervision, which improve localization of fast or blurry small objects without sacrificing efficiency.
\subsubsection{YOLOFT Demonstrates Strong Cross-Dataset Generalization}

Although recent models like D-FINE~\cite{DFINE2025} and HyperYOLO~\cite{HyperYOLO} perform well on static-image datasets, they struggle on video benchmarks due to limited temporal modeling and coarse spatial alignment. In contrast, YOLOFT-L achieves 17.8\% AP on VisDrone and ranks first in all small-object categories (AP$_{es}$ = 3.9\%, AP$_{rs}$ = 6.0\%, AP$_{gs}$ = 11.2\%), with a latency of just 29.3\,ms. On the more challenging XS-VID benchmark, YOLOFT-L achieves 29.3\% AP and sets new state-of-the-art results: 11.5\% AP$_{es}$, 16.2\% AP$_{rs}$, and 17.0\% AP$_{gs}$. These results demonstrate that YOLOFT generalizes effectively across datasets with diverse scenes, object scales, and temporal dynamics, making it a strong and efficient baseline for small-object video detection. Qualitative detector comparisons on representative XS-VID samples are provided in the SM's Sec.~III-A.

\subsection{Design Considerations for Small Object Video Detection}\label{sec:e2}

This section summarizes key insights for designing effective models tailored to small-object detection in video, based on empirical analysis from the XS-VID benchmark. We focus on three critical design dimensions: convolutional variants, upsampling strategies, and localization loss functions.

\paragraph{Convolution Variants for Representation Efficiency}
The backbone convolution operator fundamentally determines the balance between representational richness and computational cost. As reported in Tab.~\ref{tab:conv_variants}, lightweight alternatives such as DWConv reduce FLOPs (from $36.3$ G to $33.7$ G) while slightly improving overall AP ($25.2\%$ vs. $24.9\%$ baseline). More advanced dynamic designs (e.g., DYConv and ODConv) achieve comparable accuracy but provide limited benefits for small objects (AP$_{es}^{test}$ $\approx$ $9.4$–$10.2\%$). In contrast, recent innovations like FDConv and PConv yield consistent gains, reaching $25.5\%$ and $25.3\%$ AP$^{test}$. The best-performing DCNv3 achieves $25.7\%$ AP$^{test}$ and excels particularly on medium objects ($28.1\%$ AP$^m$), highlighting the advantage of adaptive spatial sampling in coping with irregular motion patterns common in videos. Importantly, these improvements come at no significant increase in model size ($\sim 13$–16 M parameters), demonstrating that architectural refinements can boost performance without prohibitive costs.
\begin{table*}[!ht]
\caption{Exploration of convolution variants for small-object video detection. All AP values are reported in \%. FLOPs are measured at input size $640^2$.}
\label{tab:conv_variants}
\centering
\setlength{\tabcolsep}{4pt}
\begin{tabular}{lrrrrrrrrrr}
\toprule
Method & AP$^{test}$ & AP$_{es}^{test}$ & AP$_{rs}^{test}$ & AP$_{gs}^{test}$ & AP$_m^{test}$ & AP$_l^{test}$ & \#Params. & FLOPs & Pub. & Year \\
\midrule
Baseline & 24.9\% & 9.5\% & 13.7\% & 14.7\% & 27.3\% & 40.2\% & 13.4 M & 36.3 G & - & - \\
DWconv~\cite{dwconv} & 25.2\% & 9.5\% & 14.3\% & 13.4\% & 27.4\% & 43.1\% & 12.4 M & 33.7 G & ICCV & 2019 \\
DYConv~\cite{dyconv} & 24.7\% & 10.2\% & 14.0\% & 14.0\% & 27.1\% & 39.6\% & 16.2 M & 33.8 G & CVPR & 2020 \\
ODConv~\cite{odconv} & 24.9\% & 9.4\% & 13.2\% & 13.7\% & 27.6\% & 41.5\% & 16.3 M & 33.8 G & ICLR & 2022 \\
FDConv~\cite{fdconv} & 25.5\% & 9.5\% & 14.1\% & 14.6\% & 27.9\% & 41.6\% & 13.5 M & 37.9 G & CVPR & 2025 \\
PConv~\cite{pconv} & 25.3\% & 9.7\% & 14.0\% & 13.8\% & 27.5\% & 43.1\% & 13.4 M & 36.2 G & AAAI & 2025 \\
DCNv3~\cite{DCNv3} & \textbf{25.7\%} & 9.7\% & 13.7\% & 14.1\% & \textbf{28.1\%} & 39.7\% & 13.6 M & 36.7 G & CVPR & 2023 \\
\bottomrule
\end{tabular}
\end{table*}

\begin{table}[htbp]
\caption{Exploration of different upsampling methods in small object detection.}
\label{tab:upsample}
\centering
\begin{tabular}{lrrrr}
\toprule
Method & AP$^{test}$ & AP$^s$ & AP$^l$ & Latency $\scriptstyle [bs=1]$ \\
 & (\%) & (\%) & (\%) & (ms) \\ 
\midrule
Bilinear & 23.9 & 12.4 & \textbf{39.1} & 20.5 \\
CARAFE~\cite{carafe} & 23.3 & 12.3 & 34.3 & 10.6 \\
ConvTranspose & 23.2 & 12.7 & 31.9 & \textbf{9.9} \\
Deconv~\cite{deconv} & 23.4 & 12.5 & 38.4 & 10.8 \\
ESPCN~\cite{espcn} & 24.5 & \textbf{13.7} & 37.1 & 11.2 \\
DySample~\cite{dysample} & \textbf{24.9} & 13.6 & 36.2 & 10.5 \\
\bottomrule
\end{tabular}
\end{table}

\paragraph{Effective Upsampling for Preserving Fine-Grained Details}  
While convolutional variants improve the efficiency and adaptability of feature representation, recovering fine-grained details still requires specialized upsampling modules. Small object detection relies heavily on high-resolution feature maps to preserve subtle spatial cues. As shown in Tab.~\ref{tab:upsample}, standard bilinear interpolation achieves competitive performance for large objects ($39.1\%$ AP$^l$) but struggles with small ones ($12.4\%$ AP$^s$). In contrast, learning-based upsampling operators such as ESPCN and DySample substantially improve AP$^s$ to $13.7\%$ and $13.6\%$, respectively, while keeping latency moderate ($\sim 11$ ms). This suggests that tailored upsampling strategies are essential for reconstructing small-scale structures lost during naive resizing. Although ConvTranspose achieves the lowest latency ($9.9$ ms), its performance on small objects ($12.7\%$ AP$^s$) lags behind, indicating that efficiency alone cannot ensure accuracy. Among the tested methods, DySample strikes the most balanced trade-off, delivering the highest overall AP ($24.9\%$) across object scales.

\begin{table}[!ht]
\caption{Exploration of small-object localization losses on XS-VID. Mixed losses are weighted combinations of IoU-based and distribution-aware terms.}
\label{tab:small_loss}
\centering
\begin{tabular}{lrrrr}
\toprule
Loss Weight & {AP}$^{test}$ & {AP}$^s$ & {AP}$^m$ & {AP}$^l$\\
 & (\%) & (\%) & (\%) & (\%)\\
\midrule
baseline (GIoU~\cite{giou}) & 23.1 & 12.0 & 24.9 & 34.7\\
1.0*CIoU & 23.2 & 12.4 & 24.5 & 33.6\\
1.0*NWD~\cite{nwdloss} & 23.2 & 12.9 & \textbf{27.2} & 35.9\\
0.4*CIoU + 0.6*NWD & 23.6 & 13.2 & 24.2 & 32.7\\
0.2*CIoU + 0.8*SAFit~\cite{RGBT-Tiny} & 23.8 & 12.8 & 25.3 &32.8 \\
0.4*CIoU + 0.6*SAFit~\cite{RGBT-Tiny} & \textbf{24.3} & \textbf{14.1} & 26.5 &\textbf{37.4} \\
\bottomrule
\end{tabular}
\end{table}

\paragraph{Localization Losses for Small Object Precision}  
Beyond feature design, accurate localization remains challenging for small objects, where slight coordinate deviations can cause large IoU penalties. Tab.~\ref{tab:small_loss} shows that replacing the baseline GIoU with distribution-aware losses such as NWD improves both small-object performance (AP$^s$ increases from $12.0\%$ to $12.9\%$) and medium-object detection ($27.2\%$ AP$^m$). More importantly, hybrid formulations that combine IoU-based and distribution-aware terms lead to the strongest results. For instance, a 0.4:0.6 mixture of CIoU and SAFit achieves the best small-object accuracy ($14.1\%$ AP$^s$), surpassing all other settings. These findings suggest that balancing overlap-sensitive and distance-aware objectives is critical for robust small object detection. Moreover, SAFit-driven hybrids maintain strong performance on large objects (AP$^l=37.4\%$), making them particularly suitable for scale-diverse video scenarios.

\paragraph{Summary of Design Insights}  
Overall, our empirical study indicates that small object video detection requires coordinated design choices across multiple components. Advanced convolution variants (e.g., FDConv~\cite{fdconv}, DCNv3~\cite{DCNv3}) enhance feature expressiveness with minimal cost; learning-based upsampling modules (e.g., DySample~\cite{dysample}, ESPCN~\cite{espcn}) are indispensable for fine detail reconstruction; and hybrid localization losses that integrate IoU- and distribution-aware terms yield superior precision on small targets. These results collectively underscore the necessity of tailoring the detection pipeline to the unique challenges of small-object perception in video.

\subsection{Ablation Study on YOLOFT}\label{sec:e3}

To systematically quantify the contribution of each architectural component within YOLOFT, we conduct ablation studies on the XS-VID benchmark. The base configuration adopts a standard YOLOv8 detector~\cite{ultralytics}, and subsequent modules are integrated incrementally. Evaluations are performed using mean Average Precision (mAP), with particular emphasis on $\text{AP}_s$ for small-object scenarios. Together with the preceding design study, these ablations assess the necessity of the small-object techniques, local temporal encoding, FMA-based aggregation, and mask-assisted supervision for the extremely small-object setting.

\subsubsection{Analysis of Modular Contributions}

Tab.~\ref{tab:ablation_components} presents the step-by-step integration of YOLOFT modules. The YOLOv8 baseline without any temporal modeling or small-object enhancements yields only $\text{AP}=22.7\%$ and $\text{AP}_s=12.6\%$. Introducing small-object specific techniques from Sec.~\ref{sec:e2}—including upsampling, adaptive backbones, and improved localization losses—boosts performance to $\text{AP}=24.5\%$, $\text{AP}_s=13.7\%$, validating their benefit in improving fine-scale feature representation.
% \vspace{-5pt}
\begin{table}[!ht]
\centering
\caption{Ablation study on the final YOLOFT design. Each row incrementally adds one component to the YOLOv8 baseline to evaluate its contribution to small, medium, and large object detection. Bold values identify the final configuration rather than column-wise maxima.}
\label{tab:ablation_components}
\setlength{\tabcolsep}{5.4pt}
\begin{tabular}{lrrrr}
\toprule
\multirow{2}{*}{Method} & AP$^{test}$ & AP$_s^{test}$ & AP$_m^{test}$ & AP$_l^{test}$ \\
 & (\%) & (\%) & (\%) & (\%) \\
\midrule
Baseline (YOLOv8) & 22.7 & 12.6 & 23.7 & 36.0 \\
+ Small object techniques & 24.5 & 13.7 & 25.8 & 37.1 \\
+ Flow Mask Attention (FMA) & 24.3 & 13.0 & 27.1 & 37.6 \\
+ Loss mask supervision & 25.2 & 14.8 & 27.4 & 43.1 \\
+ Video data augmentation & \textbf{25.8} & \textbf{14.7} & \textbf{28.1} & \textbf{39.2} \\
\bottomrule
\end{tabular}
\end{table}

Building on this, we evaluate the role of Flow Mask Attention (FMA). Directly adding FMA leads to a slight performance drop ($\text{AP}=24.3\%$, $\text{AP}_s=13.0\%$), reflecting the difficulty of learning spatiotemporal correspondence without explicit supervision. To address this, we introduce a mask-guided auxiliary loss via SAM2~\cite{ravi2024sam2}, which provides denser supervision for motion-aligned features. This addition significantly improves results to $\text{AP}=25.2\%$ and $\text{AP}_s=14.8\%$.

Finally, incorporating video-level augmentations enhances robustness under motion and viewpoint shifts, pushing accuracy to $\text{AP}=25.8\%$, $\text{AP}_m=28.1\%$, and $\text{AP}_s=14.7\%$. As shown in Tab.~\ref{tab:data_argument}, moderate transformations (e.g., small-degree rotation and scaling) yield improvements, whereas overly aggressive distortions can harm the structural integrity of small objects.

\begin{table}[!ht]
\centering
\caption{Effect of different data augmentation strategies on the XS-VID snapshot (D: Degree, S: Scale, P: Perspective). Reproducibility details are provided in the SM's Sec.~II-F.}
\label{tab:data_argument}
\small
\begin{tabular}{lrrrr}
\toprule
{[D, S, P]} & AP$^{test}$ & AP$^{es}$ & AP$^{rs}$ & AP$^{gs}$ \\
            & (\%) & (\%) & (\%) & (\%) \\
\midrule          
{[2.0, 0.2, 1e-4]}   & \textbf{32.6} & \textbf{21.3} & 26.2 & \textbf{33.6} \\
No augmentation      & 32.3          & 20.0          & \textbf{26.4} & 33.2 \\
{[0.0, 0.5, 1e-3]}   & 32.1          & 20.8          & 25.3 & 33.1 \\
{[0.0, 0.0, 1e-3]}   & 31.8          & 20.6          & 26.4 & 32.6 \\
\bottomrule
\end{tabular}
\end{table}

\subsubsection{Ablation on Temporal Fusion Strategy}

To isolate the effects of different temporal fusion strategies within FMA, we evaluate five configurations: naive concatenation, element-wise addition, ConvGRU, ConvLSTM, and 3D convolution. Results are summarized in Tab.~\ref{tab:temporal_ablation}.

Concatenation and addition yield limited improvements over the single-frame baseline (23.3\% and 23.6\% mAP, respectively), demonstrating that rudimentary fusion lacks sufficient temporal context encoding capacity. ConvGRU improves performance to 23.9\%, with $\text{AP}_s=12.8\%$, validating the utility of selective temporal memory. However, ConvLSTM slightly underperforms (23.5\%), potentially due to training instability and gradient vanishing in small-object contexts.

While recurrent units introduce temporal gating mechanisms, their latency (15–18 ms per frame) and parameter cost are nontrivial. By contrast, our 3D convolution approach—applied on stacked frame features—achieves the best accuracy (mAP = 24.7\%, $\text{AP}_s=13.4\%$), with reduced computational burden (23.7G FLOPs, 13.9 ms latency). These findings indicate that localized spatio-temporal filters learned by 3D convolutions provide more efficient and expressive temporal integration under computational constraints.
\begin{table}[!ht]
\centering
\caption{Ablation study on different temporal modeling strategies. FLOPs are measured at input size $640^2$, and latency is tested with batch size 1 under FP16 precision.}
\label{tab:temporal_ablation}
\begin{tabular}{lrrrr}
\toprule
\multirow{2}{*}{Method} & AP$^{test}$ & AP$^s$ & FLOPs & Latency \\
 & (\%) & (\%) & (G) & (ms) \\
\midrule
Concat & 23.3 & 12.5 & 29.1 & 10.0 \\
Add & 23.6 & 12.3 & 28.8 & 9.0 \\
ConvGRU & 23.9 & 12.8 & 29.3 & 15.2 \\
LSTM & 23.5 & 12.4 & 31.7 & 18.4 \\
3D Conv & \textbf{24.7} & \textbf{13.4} & \textbf{23.7} & \textbf{13.9} \\
\bottomrule
\end{tabular}
\end{table}

\subsubsection{Ablation on Mask Supervision Strategy}

We further examine the role of auxiliary mask supervision, particularly how the temporal scope of supervision affects detection quality. Four variants are considered: (i) no mask loss, (ii) current-frame mask only, (iii) full-weight supervision from both current and previous frame masks, and (iv) half-weight supervision from the historical mask.

As shown in Tab.~\ref{tab:loss_mask_ablation}, using current-frame masks alone improves mAP from 23.2\% to 24.1\%, validating the utility of pixel-wise loss in regions aligned with ground-truth boxes. Adding unweighted historical supervision (weight = 1) slightly degrades performance (23.5\%), likely due to misalignment noise in pseudo-mask propagation. By contrast, applying a soft-weighted historical mask (weight = 0.5) yields the best result (24.6\%), highlighting the benefit of temporal consistency when appropriately regularized.
\begin{table}[!ht]
\centering
\caption{Ablation study on mask-assisted supervision. We vary whether historical and current frame masks are used, and the weight assigned to the auxiliary loss.}
\label{tab:loss_mask_ablation}
\begin{tabular}{cccc}
\toprule
\multirow{2}{*}{History} & \multirow{2}{*}{Current} & \multirow{2}{*}{Weight} & AP$^{test}$ (\%) \\
 & & & \\
\midrule
-- & -- & -- & 23.2 \\
-- & \checkmark & 1.0 & 24.1 \\
\checkmark & \checkmark & 1.0 & 23.5 \\
\checkmark & \checkmark & 0.5 & \textbf{24.6} \\
\bottomrule
\end{tabular}
\end{table}

Across all configurations, the largest $\Delta \text{AP}_s$ is observed with mask supervision (+2.2\%), indicating that pixel-level attention priors strongly enhance small-object learning. Additionally, the improved convergence of the FMA module under this supervision suggests that the auxiliary task effectively anchors temporal dynamics during training.

\section{Conclusion}

We introduce XS-VID, a large-scale video benchmark designed to advance the study of small-object detection and tracking. With over 1.4 million annotations spanning 16 diverse scene types and a wide range of object scales, XS-VID provides extensive coverage of extremely small instances—29.4\% of all objects are smaller than $12^2$ pixels. It supports three evaluation tracks: Detection, SOT, and MOT, enabling detailed analysis under realistic video conditions. Compared to existing datasets, XS-VID offers greater scale, object diversity, and task coverage for benchmarking small-object performance.

Based on insights from this benchmark, we propose YOLOFT, a lightweight video detection framework that integrates flow-guided temporal modeling and mask-based auxiliary supervision. These modules improve spatial alignment and temporal consistency, which are critical for detecting small, fast-moving, and low-texture objects. YOLOFT achieves 29.3\% mAP(+1.0\%) on XS-VID and best AP$_{es}$(+2.3\%) on VisDrone, setting new baselines for real-time small-object video detection.

XS-VID and YOLOFT together provide a solid foundation for future research. Extensions to longer-term temporal modeling or multi-modal fusion may further improve performance in complex scenarios.

\section*{Acknowledgments}
This work was supported in part by the National Natural Science Foundation of China (NSFC) under Grant 62276108.

\bibliographystyle{IEEEtran}
\bibliography{bibliography/ref}
\begin{IEEEbiography}
[{\includegraphics[width=1in,height=1.25in,clip,keepaspectratio]{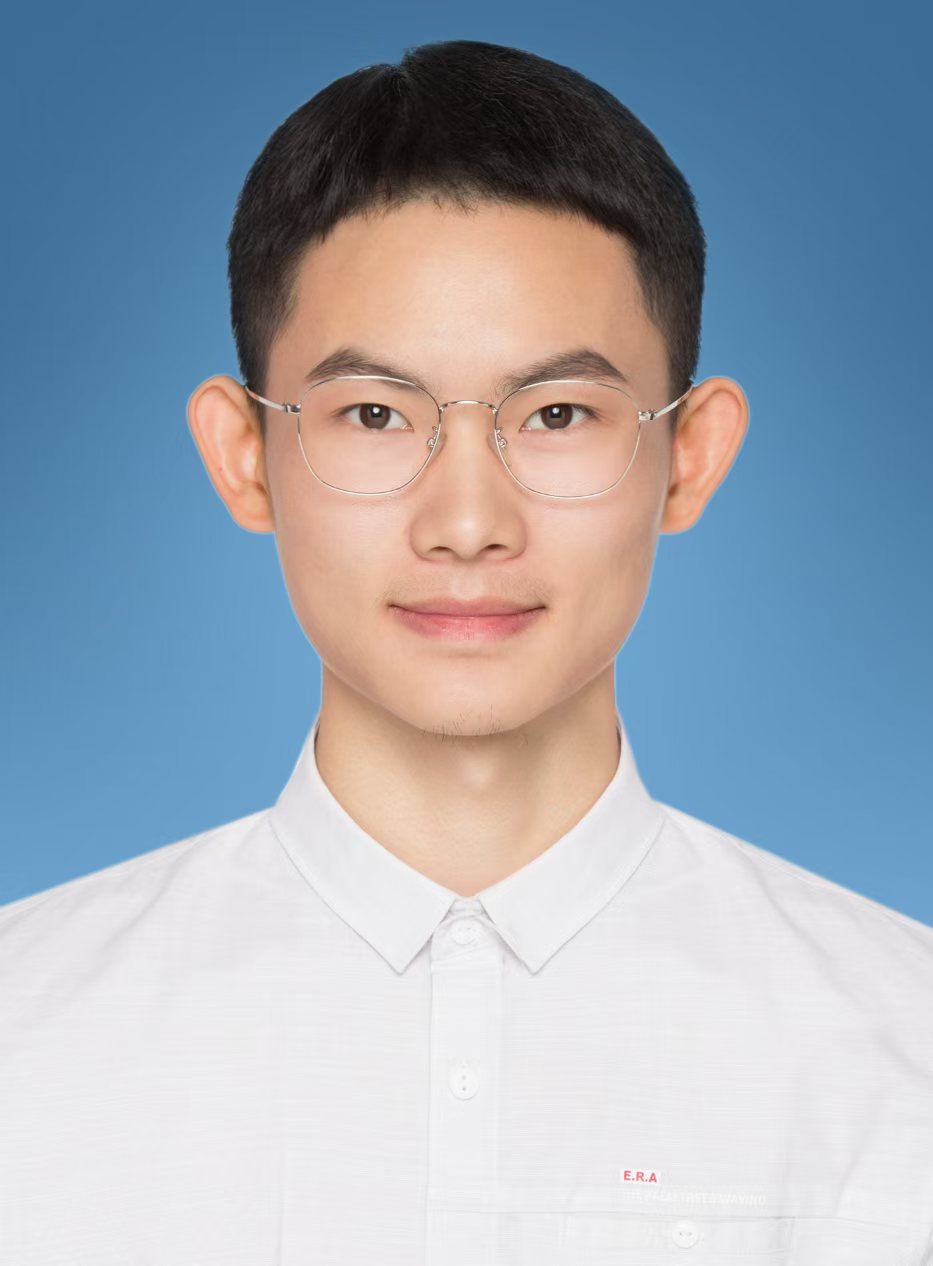}}]{Jiahao Guo} received his B.E. degree in Electronic Information Engineering from the School of Electronic Information and Communications, Huazhong University of Science and Technology, China, in 2024. He is currently pursuing an M.E. degree at the same school. His research interests include object detection, language models, and artificial intelligence.
\end{IEEEbiography}
\vspace{-50pt}
\begin{IEEEbiography}[{\includegraphics[width=1in,height=1.25in,clip,keepaspectratio]{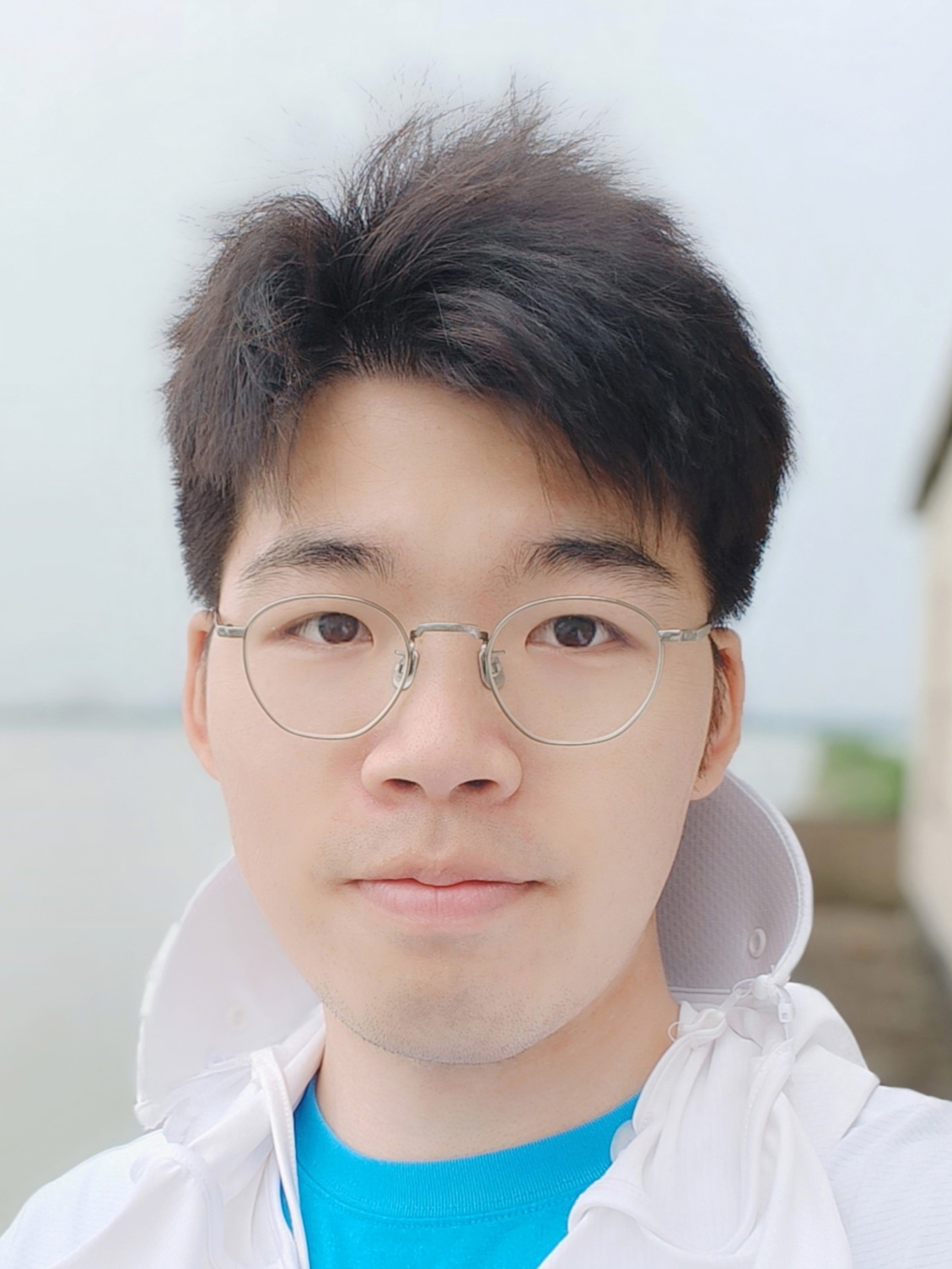}}]{Ziyang Xu} received his M.E. degree in Information and Communication Engineering from Huazhong University of Science and Technology (HUST), China, in 2025. He is currently pursuing a Ph.D. degree at the School of EIC, HUST. He has published several peer-reviewed top conference and journal papers, including Nature Medicine, Med (Cell Press), NeurIPS, AAAI, ECCV, and ECAI. His current research interests lie in generative AI, with a focus on visual generation, perception, and memory.
\end{IEEEbiography}
\vspace{-50pt}
\begin{IEEEbiography}
[{\includegraphics[width=1in,height=1.25in,clip,keepaspectratio]{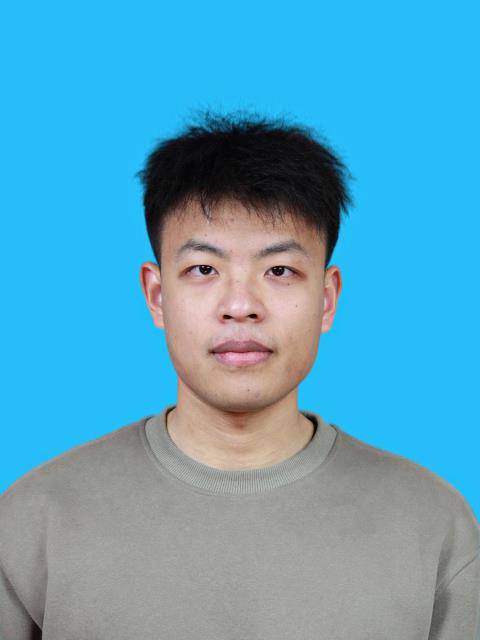}}]{Lianjun Wu} received his B.E. degree in Electronic Information Engineering from the School of Electronic Information and Communications, Huazhong University of Science and Technology, China, in 2022, and the M.E. degree from the same school in 2025. His research interests include object detection and segmentation, medical image processing, and artificial intelligence. He has authored peer-reviewed publications in top-tier conferences and journals, including Image and Vision Computing and ECCV.
\end{IEEEbiography}
\vspace{-50pt}
\begin{IEEEbiography}
[{\includegraphics[width=1in,height=1.25in,clip,keepaspectratio]{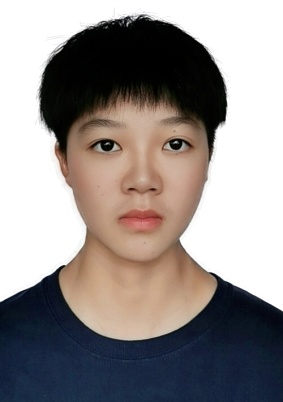}}]{Fei Gao} received her B.E.degree in Information Engineering from Wuhan University of Technology, China, in 2023. She is currently pursuing an M.E. degree at the School of Electronic Information and Communications, Huazhong University of Science and Technology. Her research interests include video frame Interpolation, video reconstruction and artiﬁcial intelligence.
\end{IEEEbiography}
% \newpage
\vspace{-50pt}
\begin{IEEEbiography}[{\includegraphics[width=1in,height=1.25in,clip,keepaspectratio]{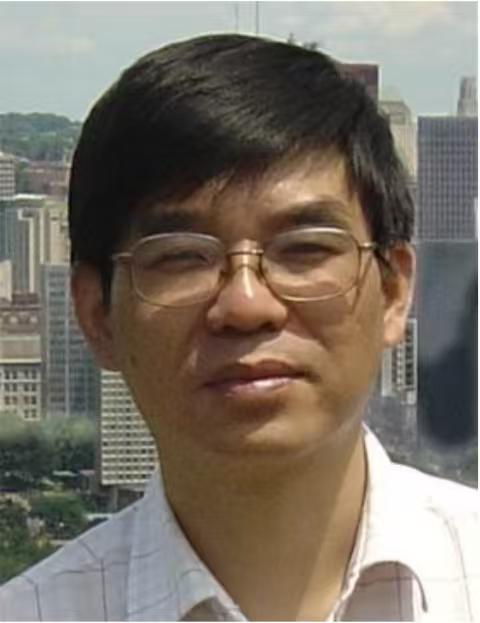}}]{Wenyu Liu (SM'15)}
    received the B.S. degree in Computer Science from Tsinghua University, Beijing, China, in 1986, and the M.S. and Ph.D. degrees, both in Electronics and Information Engineering, from Huazhong University of Science and Technology (HUST), Wuhan, China, in 1991 and 2001, respectively. He is now a professor and associate dean of the School of Electronic Information and Communications, HUST. His research areas include computer vision, multimedia, and machine learning.
\end{IEEEbiography}
\vspace{-50pt}
% Cancel the stretchable space inserted before the next biography.
% \vskip 0pt plus -1fil
% % Now this value controls the actual fixed gap.
% \vspace{-15pt}
\begin{IEEEbiography}[{\includegraphics[width=1in,height=1.25in,clip,keepaspectratio]{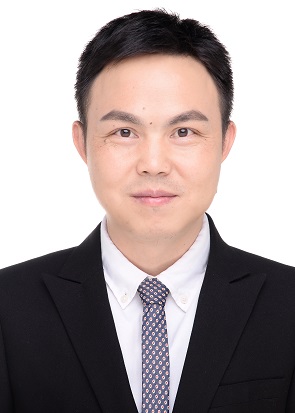}}]{Xinggang Wang}
  (SM'24) received his B.S. and Ph.D. degrees in Electronics and Information Engineering from Huazhong University of Science and Technology (HUST), Wuhan, China, in 2009 and 2014, respectively. He was a visiting scholar at Temple University and UCLA. Currently, he serves as a Professor at the School of Electronic Information and Communications at HUST. His research interests encompass visual representation learning, visual perception, visual planning, and multimodal foundation models. He has over 60,000 Google Scholar citations and an h-index of 90. He is the Co-Editor-in-Chief of Image and Vision Computing, an Associate Editor for IEEE TPAMI and Machine Vision Applications, and has served as an Area Chair for CVPR, ICCV, and NeurIPS.
\end{IEEEbiography}

\clearpage
\twocolumn[{
\begin{center}
  {\LARGE\bfseries Supplementary Material of XS-VID\par}
  \vspace{0.6em}
  {\large Jiahao Guo, Ziyang Xu, Lianjun Wu, Fei Gao, Wenyu Liu, Xinggang Wang\par}
\end{center}
\vspace{1em}
}]

\setcounter{section}{0}
\setcounter{table}{0}
\setcounter{figure}{0}
\renewcommand{\thetable}{S\arabic{table}}
\renewcommand{\thefigure}{S\arabic{figure}}
\renewcommand{\theHsection}{supplement.\arabic{section}}
\renewcommand{\theHsubsection}{supplement.\arabic{section}.\arabic{subsection}}
\renewcommand{\theHtable}{supplement.\arabic{table}}
\renewcommand{\theHfigure}{supplement.\arabic{figure}}

\section{XS-VID Data Construction and Annotation}\label{sec:Data_Details}

\subsection{Data Collection and Resolution Standardization}
We used DJI Air3 drones flying at altitudes of 70–90m and speeds of 5–10m/s, capturing 500 videos at $3840 \times 2160$ resolution and 25fps. Scenarios include urban streets, highways, riversides, and wooded areas, across different seasons and lighting conditions. After quality screening and scene filtering, 204 self-collected videos were retained. An additional 170 drone or elevated-view videos were collected from online sources, yielding 374 sequences in total.

\textbf{Post-processing.} Following high-resolution annotation, frames are converted to an intermediate $1920\times1080$ representation, divided into overlapping $1080\times1080$ square units, and bilinearly resized to the released $1024\times1024$ benchmark samples. The original high-resolution videos and annotations will also be released after licensing and privacy screening to support native-resolution and alternative-processing research.

\subsection{Annotation Protocol and Quality Assurance}

\textbf{Classification.} The dataset includes seven classes: person, car, bicycle, cyclist, boat, truck, and ignore (regions that are hard to annotate yet remain detection targets are labeled "ignore" to aid researchers and avoid interfering with model training. During training, these ignore regions are excluded from background supervision). Extremely small objects like pedestrians and bicycles often suffer from occlusion and low resolution, complicating annotation.

\textbf{Annotation Pipeline.}
\begin{itemize}
  \item[1)] Use CO-DETR~\cite{codetr2022} for pre-annotation.
  \item[2)] Manually refine at 5–10 frame intervals.
  \item[3)] Use SIFT and FLANN~\cite{lowe2004distinctiveSIFT, muja2009FLANN} to assign consistent IDs and interpolate.
  \item[4)] Perform manual frame-by-frame checks, adjusting for missed/false annotations and occlusions.
  \item[5)] Cross-verify with 2–3 annotators per video.
\end{itemize}

\textbf{Annotation Guidelines.}
\begin{itemize}
  \item Objects with >50\% occlusion or strong overlap are not annotated.
  \item Clustered pedestrians or multiple-object regions are labeled as ignore.
  \item Ads, reflections, and ambiguous shapes are consistently excluded.
  \item Ignore labels are only applied when visual ambiguity cannot be resolved, even by models.
\end{itemize}

\textbf{Quality Assurance.} Each video is double- or triple-annotated. On test samples, we observed an inter-annotator IoU exceeding 0.85. Object lifespan across frames was verified for consistency.

\textbf{Annotation Efficiency.} This hybrid pipeline reduces manual workload by over 50\%, while maintaining high fidelity, especially for extremely small instances.

\textbf{Pre-annotation--annotation comparison.} We compare CO-DETR pre-annotations (score $\geq0.3$) with the final XS-VID annotations at IoU $\geq0.5$. The initial proposals match 24\% of the released extremely small boxes, while the remaining 76\% are recovered during manual annotation.

\subsection{Data Splits}
We selected 66 fully annotated videos as the test set, ensuring that their shooting locations do not appear in the remaining 308 videos. Additionally, these 66 videos include scenes not present in the training set. This split is intended to evaluate transfer to locations and scenes absent from training.  For model development, we designate 16 videos with diverse scene characteristics (31,024 frames and 119,710 boxes) from the original 308-video training partition as a validation subset. This leaves 292 videos for training during development. After model selection, models used for final benchmark reporting are trained on the complete 308-video training partition and evaluated on the 66-video benchmark test set. See Tab.~\ref{tab:Data_split} for details.

\begin{table}[htbp]
  \centering
  \caption{Data split for XS-VID.}
  \scriptsize
  \setlength{\tabcolsep}{3pt}
  \begin{tabular}{@{}lrrr@{}}
    \toprule
    Split & \shortstack{\\\# of Sequences} & \# of Images & \shortstack{\\\# of Box Annotations} \\
    \midrule
     &  &  &  \\
    Development train & 292 & 155,422 & 929,890 \\
    Development validation & 16 & 31,024 & 119,710 \\
    Benchmark test & 66 & 36,478 & 370 K \\
    \midrule
    Total & 374 & 222924 & 1,419 K \\
    \bottomrule
  \end{tabular}%
  \label{tab:Data_split}%
\end{table}%

\begin{table}[h]
  \centering
  \caption{Method-group summary of evaluation protocols for the added baselines.}
  \label{tab:protocol_summary}
  \scriptsize
  \setlength{\tabcolsep}{3pt}
  \begin{tabular}{p{0.26\linewidth}p{0.66\linewidth}}
    \toprule
    Method group & Protocol \\
    \midrule
    Trainable baselines & Fine-tuning on XS-VID under the corresponding official setting. \\
    Added SOT trackers & XS-VID adaptation under the corresponding official task protocol. \\
    MOT tracking-by-detection & Detector outputs come from the detector named in the table; association follows the corresponding tracker implementation. \\
    Unified frameworks & Staged task adaptation following established unified-model practice~\cite{Unicorn,UNINEXT}. \\
    \bottomrule
  \end{tabular}
\end{table}
\section{Evaluation Protocols and Reproducibility}\label{sec:ExperimentalSetup}

\subsection{Common Experimental Setup}
For all GOD methods and YOLOX, we used mmdetection~\cite{mmdetection} for testing. For some VOD methods (~\cite{Zhu_2017_CVPRdff,Zhu_2017_ICCVfgfa,wu2019selsa,TROI_2021}), we used mmtracking~\cite{mmtrack2020}. Some YOLO methods were tested using Ultralytics~\cite{ultralytics}. Other methods were tested using the official code provided by the original papers. During testing, we only modified the dataset and set the training and testing image size to  $1024\times1024$, faithfully following the configurations in the original papers. Except for methods explicitly labeled as zero-shot, trainable baselines were fine-tuned on XS-VID following their corresponding official settings. Our machines are four NVIDIA RTX 3090s.

\begin{table*}[!t]
    \centering
    \renewcommand{\arraystretch}{1.08}
    \caption{Complete MOT evaluation on XS-VID. Multi-stage methods use YOLOFT detections and list the tracker and association paradigm separately; end-to-end and unified methods are reported as model-level entries. T-by-D, GNN, and E2E-Trans denote tracking by detection, graph neural network association, and end-to-end Transformer tracking, respectively.}
    \label{tab:mot_results_full}
    \footnotesize
    \setlength{\tabcolsep}{2.6pt}
    \begin{tabular}{ll l *{5}{r} r r r}
        \toprule
        \multirow{2}{*}{Detector} &
        \multirow{2}{*}{Tracker / model} &
        \multirow{2}{*}{Paradigm} &
        \multicolumn{5}{c}{TAO Tracking mAP (\%)} &
        \multicolumn{3}{c}{MOTChallenge} \\
        \cmidrule(lr){4-8}\cmidrule(lr){9-11}
        & & & mAP & AP$_s$ & AP$_m$ & AP$_s^*$ & AP$_m^*$ & MOTA (\%) & IDF1 (\%) & IDSw \\
        \midrule
        \multicolumn{11}{c}{\textit{Multi-stage}} \\
        \midrule
        YOLOFT & SUSHI~\cite{SUSHI} & GNN & 15.1 & 1.4 & 15.1 & 9.7 & 27.5 & 26.7 & 40.5 & 4821 \\
        YOLOFT & BoostTrack~\cite{BoostTrack} & T-by-D & 14.4 & 8.0 & 11.6 & 9.9 & 14.6 & 17.3 & 40.3 & 9616 \\
        YOLOFT & HybridSORT~\cite{HybridSORT} & T-by-D & 16.5 & 4.0 & 16.2 & 11.9 & 26.6 & 37.5 & 48.7 & 2076 \\
        YOLOFT & OC-SORT~\cite{ocsort} & T-by-D & 20.1 & 8.8 & 26.1 & 11.6 & 32.0 & 34.8 & 47.8 & 7116 \\
        YOLOFT & ByteTrack~\cite{ByteTrack} & T-by-D & 22.3 & 11.4 & 26.7 & 11.9 & 34.3 & 36.1 & 52.6 & 1603 \\
        YOLOFT & StrongSORT~\cite{Strongsort} & T-by-D & 26.0 & 18.9 & 25.8 & 13.9 & 38.3 & 36.8 & 53.9 & 4283 \\
        YOLOFT & DeepOCSORT~\cite{deepocsort} & T-by-D & 26.9 & 20.0 & 25.2 & 16.6 & 39.1 & 37.4 & 53.0 & 3563 \\
        YOLOFT & BoT-SORT~\cite{botsort} & T-by-D & \textbf{32.1} & \textbf{22.4} & \textbf{33.4} & \textbf{18.1} & \textbf{40.8} & 37.1 & \textbf{54.7} & 3411 \\
        \midrule
        \multicolumn{11}{c}{\textit{End-to-end and unified}} \\
        \midrule
        \multicolumn{2}{c}{TrackFormer~\cite{TrackFormer}} & E2E-Trans & 15.2 & 3.7 & 18.5 & 1.5 & 26.2 & 27.6 & 41.2 & 1680 \\
        \multicolumn{2}{c}{MOTR~\cite{MOTR}} & E2E-Trans & 13.0 & 2.8 & 13.4 & 10.3 & 17.3 & 19.8 & 35.7 & 65 \\
        \multicolumn{2}{c}{MOTIP~\cite{MOTIP}} & E2E-Trans & 6.5 & 0.5 & 7.6 & 0.0 & 20.3 & 13.2 & 36.2 & 2549 \\
        \multicolumn{2}{c}{Unicorn~\cite{Unicorn}} & Unified & 24.6 & 10.2 & 31.2 & 6.4 & 33.4 & 32.4 & 51.2 & 1537 \\
        \multicolumn{2}{c}{UNINEXT~\cite{UNINEXT}} & Unified & 20.1 & 11.5 & 23.5 & 3.3 & 29.4 & 32.6 & 50.2 & 1757 \\
        \multicolumn{2}{c}{YOLOFT} & Unified & \textbf{31.5} & \textbf{21.7} & \textbf{32.8} & \textbf{16.4} & \textbf{40.4} & 30.7 & \textbf{55.2} & 1220 \\
        \bottomrule
    \end{tabular}
\end{table*}

\textbf{Reproducibility.} The 16-video validation subset is intended for model development. After model selection, models used for final benchmark reporting are trained on the complete 308-video training partition. The train/test lists, annotations, evaluated-category and ignore-region definitions, $1024\times1024$ preprocessing, method configurations, and evaluator will be released together as one reproduction package.

\subsection{Detection Track Protocol}

Image-based and temporal detection methods use the same frame-level evaluator and the unchanged XS-VID annotations. ZS denotes zero-shot evaluation, and FT denotes fine-tuning on XS-VID. Other trainable baselines follow their corresponding official settings.

\subsection{SOT Protocol}

The SOT track uses the same 308-video training partition as the Detection Track for training or adaptation, and its test sequences are extracted only from the 66 scene-disjoint test videos. Up to five of the longest eligible trajectories are selected from each test video after excluding trajectories dominated by prolonged full occlusion. This procedure yields 272 target-specific sequences comprising 107,788 evaluated target-frame instances. Added trainable trackers follow their corresponding official task protocols. Evaluation follows one-pass evaluation with AUC, 20-pixel precision, and speed in frames per second.

\subsection{MOT Protocol}

AP$_s$/AP$_m$ use average trajectory box-area ranges of $(0,32^2]$ and $(32^2,96^2]$ pixels$^2$, respectively. AP$_s^*$/AP$_m^*$ use duration ranges of $(0,25]$ and $(25,250]$ frames, respectively; trajectories longer than 250 frames lie outside this duration-based subset. The tracking-by-detection baselines share YOLOFT detections and a primary threshold of 0.1, retaining detections down to 0.001 where low-score association applies. MOTA, IDF1, and IDSw use a common output threshold of 0.3. Remaining tracker settings and all other MOT methods follow their official recipes.

\subsection{Unified YOLOFT Extension}

Drawing on the staged training paradigm of unified models~\cite{Unicorn,UNINEXT}, we initialize from the Detection-trained YOLOFT-L and freeze its small-object temporal backbone when adding task-specific SOT and MOT heads. The SOT head follows TransT~\cite{TransT}, while the MOT head follows quasi-dense similarity learning~\cite{qdtrack}. For each task, only the corresponding head is fine-tuned for 10 epochs, retaining the small-object temporal prior learned during Detection training. Let $F'_z$ and $F'_t$ denote the template and current representations used by the corresponding heads:
\begin{equation}
\begin{aligned}
H_t^{\mathrm{SOT}} &= \mathcal{T}(F'_z,F'_t),\\
(s_t,b_t) &= \bigl(\mathcal{C}(H_t^{\mathrm{SOT}}),\,\sigma(\mathcal{B}(H_t^{\mathrm{SOT}}))\bigr),\\
E_t^{\mathrm{MOT}} &= \operatorname{Norm}\!\left(\phi(F'_t)\right).
\end{aligned}
\end{equation}
Here, $\mathcal{T}$ denotes template--current cross-attention. The operators $\mathcal{C}$ and $\mathcal{B}$ are the TransT classification and regression predictors, respectively~\cite{TransT}. The map $\phi$ produces normalized MOT instance embeddings under quasi-dense similarity supervision~\cite{qdtrack}.

\subsection{XS-VID Snapshot Reproducibility}

\noindent{\textbf{Training video IDs (26; \texttt{videos.id}):} \texttt{282--307}.}\par
\medskip
\noindent{\textbf{Test video IDs (12; \texttt{videos.id}):} \texttt{54--65}.}\par

\section{Additional Analyses and Ablations}\label{sec:AdditionalAnalysis}

\subsection{Qualitative Detector Comparison}\label{sec:supp_model_pred}

\begin{figure*}[!t]
    \centering
    \includegraphics[width=1.0\linewidth]{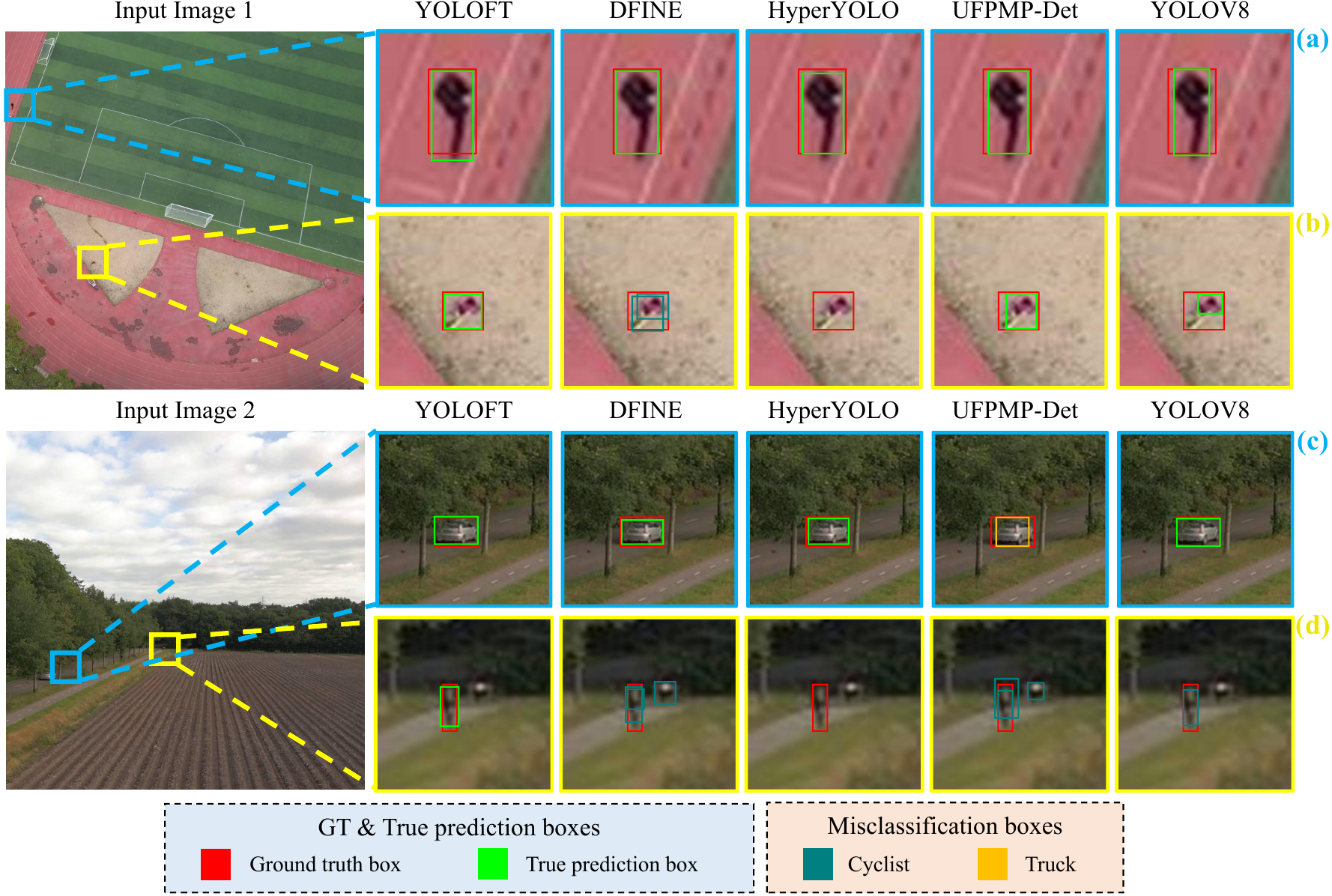}
    \caption{\textbf{Qualitative comparison of YOLOFT with baseline detectors.} YOLOFT produces predictions that align closely with ground-truth (GT) boxes across diverse scenarios. In case (a), YOLOFT captures the target size and position in a challenging small-object example. In case (c), it also detects medium-sized objects, indicating performance across object scales.}
    \label{fig:model_pred}
    \vspace{-10pt}
\end{figure*}

Fig.~\ref{fig:model_pred} compares YOLOFT with YOLOv8, DFINE, HyperYOLO, and UFPMP-Det on representative XS-VID samples. In challenging cases such as (b) and (c), the baseline detectors can misclassify or miss extremely small objects in complex backgrounds. YOLOFT reduces category confusion and produces boxes that more closely align with the GT annotations. The medium- and large-object case in (c) further indicates that its behavior is not limited to the smallest scale range.

\subsection{Resolution Analysis}

We use the official CO-DETR pre-annotation model to compare high- and low-resolution pre-annotation on the self-collected 4K subset. The high-resolution branch preserves source detail through square tiling; the low-resolution branch downsamples the whole source frame before the same square input interface. Both branches use the same ground truth for the released tiles and the same class-agnostic box-scoring protocol. High-resolution pre-annotation obtains 10.9\% AP, compared with 2.2\% for low-resolution pre-annotation (Tab.~\ref{tab:preanno_resolution}). This gap supports retaining high-resolution evidence during pre-annotation and human correction.

\begin{table}[htbp]
  \centering
  \caption{High- versus low-resolution pre-annotation using the official CO-DETR checkpoint.}
  \label{tab:preanno_resolution}
  \scriptsize
  \setlength{\tabcolsep}{3.5pt}
  \resizebox{\columnwidth}{!}{%
  \begin{tabular}{lrrrrrr}
    \toprule
    Pre-annotation input & Box AP (\%) & AP$_{es}$ (\%) & AP$_{rs}$ (\%) & AP$_{gs}$ (\%) & AP$_m$ (\%) & AP$_l$ (\%) \\
    \midrule
    High resolution & \textbf{10.9} & \textbf{3.0} & \textbf{9.5} & \textbf{6.6} & \textbf{30.5} & \textbf{35.2} \\
    Low resolution & 2.2 & 0.0 & 0.1 & 0.5 & 9.4 & 10.1 \\
    \bottomrule
  \end{tabular}
  }
\end{table}

\subsection{Ignore-Region Ablation}

The reported YOLOFT-L configuration uses the benchmark's ignore-region handling and obtains 29.3\% AP and 11.5\% AP$_{es}$. In matched training without ignore handling, the results are 28.6\% AP and 11.3\% AP$_{es}$ (Tab.~\ref{tab:ignore_ablation}). In this ablation, ignore-region handling is associated with gains of 0.7 AP and 0.2 AP$_{es}$.

\begin{table}[htbp]
  \centering
  \caption{Effect of ignore-region handling on YOLOFT-L training. Values in parentheses denote changes relative to the setting with ignore-region handling.}
  \label{tab:ignore_ablation}
  \small
  \begin{tabular}{lcc}
    \toprule
    Setting & AP (\%) & AP$_{es}$ (\%) \\
    \midrule
    With ignore & 29.3 & 11.5 \\
    Without ignore & 28.6 ($-0.7$) & 11.3 ($-0.2$) \\
    \bottomrule
  \end{tabular}
\end{table}

\begin{figure*}[h]
    \centering
    \subfloat[Example set 1\label{fig:dataset_show1}]{\includegraphics[height=0.48\textheight]{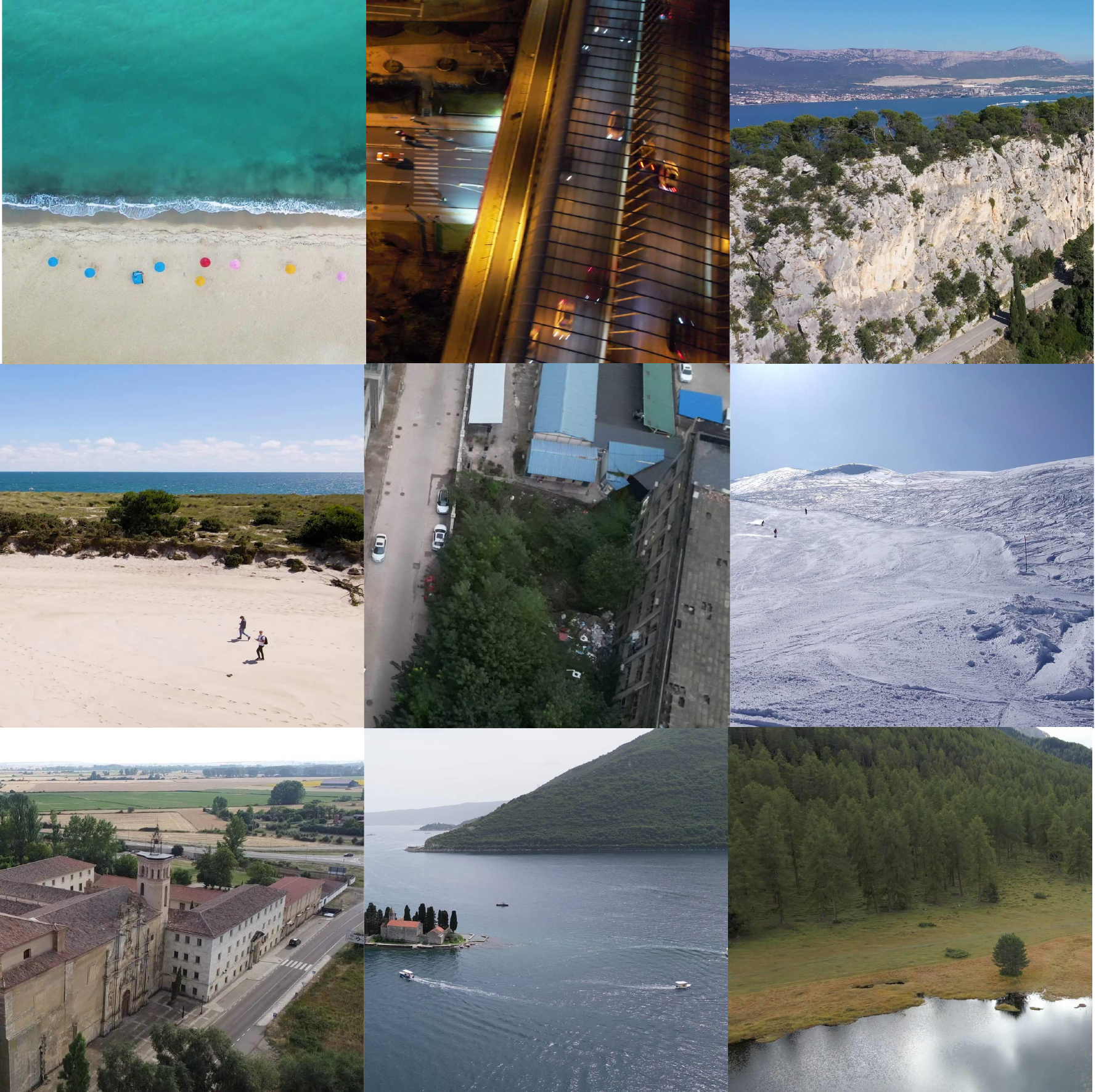}}
    \hspace{0.04\textwidth}
    \subfloat[Example set 2\label{fig:dataset_show2}]{\includegraphics[height=0.48\textheight]{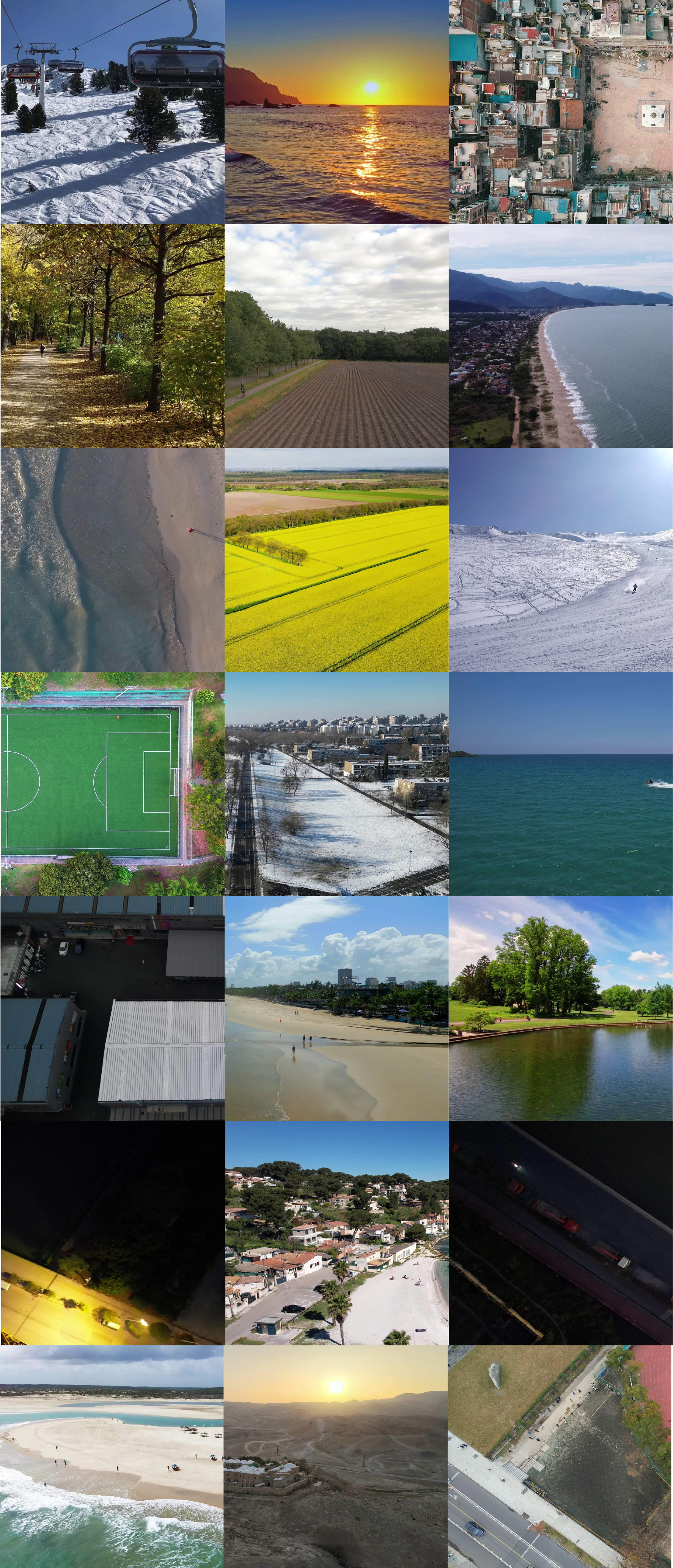}}
    \caption{Representative XS-VID scenes illustrating variation in environment, viewpoint, illumination, target density, and extremely small-object appearance.}
    \label{fig:dataset_examples}
\end{figure*}

\section{Limitations}\label{sec:Limitations}
The bounding box annotations, despite multiple rounds of cross-validation, remain subject to inherent human annotation challenges. Subtle ambiguities in object boundaries—such as partially occluded or irregularly shaped objects (e.g., overlapping crowds)—may cause inconsistent bounding box placements. Additionally, defining the first/last frames for objects entering/exiting drone-captured frames is inherently ambiguous, leading to incomplete track annotations, especially for fast-moving targets. Our baseline method YOLOFT also has limitations in handling complex motions. Its local streaming aggregation may not capture global motion patterns of larger objects (e.g., trucks), for which we observe occasional localization and classification errors.

\section{Potential Negative Societal Impacts}\label{sec:Negative}
This dataset, including drone-captured videos and web-crawled data, carries potential risks, mitigated as follows:
\begin{itemize}
\item \textbf{Privacy}: Drone footage has sensitive content obscured; web-crawled data is filtered to remove visible personal identifiers where feasible, and release materials will be reviewed under the applicable privacy requirements.
\item \textbf{Security}: Drone footage and web data are screened to reduce the risk of releasing sensitive locations or content that could enable harmful use.
\item \textbf{Bias and Usage}: Geographic/category imbalances will be addressed by expanding data diversity. All data is restricted to academic research use only.
\end{itemize}

\section{Visualization cases of XS-VID Dataset} \label{sec:showdataset_cases}

% \begin{figure*}
%     \centering
%     \includegraphics[width=1.0\linewidth]{figures/main/dataset show.pdf}
%     \caption{\textbf{Visualization cases of XS-VID Dataset.}}
%     \label{fig:dataset_show}
% \end{figure*}

As shown in Fig.~\ref{fig:dataset_examples}, the XS-VID examples span roads, forests, buildings, bridges, and both daytime and nighttime conditions.

\end{document}